\documentclass[final]{elsarticle}

\usepackage{framed,multirow}

\usepackage[breaklinks=true]{hyperref}

\usepackage{graphicx,subcaption}

\usepackage{array}
\usepackage{tabularx} 
\usepackage[numbers]{natbib}
\usepackage{adjustbox}
\usepackage{amsmath,bm}
\usepackage{amssymb}
\usepackage{siunitx}
\usepackage{textcomp}
\usepackage{mathtools}
\usepackage{scalerel}
\usepackage{tabularx,multirow,makecell}
\usepackage[ruled,vlined]{algorithm2e}
\usepackage{setspace}
\usepackage{xcolor}
\usepackage{color}
\usepackage{colortbl}

\usepackage{float}
\usepackage{placeins}
\usepackage{listings}

\colorlet{punct}{red!60!black}
\definecolor{background}{HTML}{EEEEEE}
\definecolor{delim}{RGB}{20,105,176}
\colorlet{numb}{magenta!60!black}

\lstdefinelanguage{json}{
    basicstyle=\normalfont\ttfamily,
    numbers=left,
    numberstyle=\scriptsize,
    stepnumber=1,
    numbersep=8pt,
    showstringspaces=false,
    breaklines=true,
    frame=lines,
    backgroundcolor=\color{background},
    literate=
     *{0}{{{\color{numb}0}}}{1}
      {1}{{{\color{numb}1}}}{1}
      {2}{{{\color{numb}2}}}{1}
      {3}{{{\color{numb}3}}}{1}
      {4}{{{\color{numb}4}}}{1}
      {5}{{{\color{numb}5}}}{1}
      {6}{{{\color{numb}6}}}{1}
      {7}{{{\color{numb}7}}}{1}
      {8}{{{\color{numb}8}}}{1}
      {9}{{{\color{numb}9}}}{1}
      {:}{{{\color{punct}{:}}}}{1}
      {,}{{{\color{punct}{,}}}}{1}
      {\{}{{{\color{delim}{\{}}}}{1}
      {\}}{{{\color{delim}{\}}}}}{1}
      {[}{{{\color{delim}{[}}}}{1}
      {]}{{{\color{delim}{]}}}}{1},
}

\usepackage{pifont}

\colorlet{tableheadcolor}{gray!33} %

\usepackage{booktabs}
\usepackage{rotating}

\usepackage[flushleft]{threeparttable}

\biboptions{authoryear}

\usepackage{setspace}
\usepackage{xcolor}
\usepackage{color}
\usepackage{soul}

\def\XS{\xspace}
\def\figureabvr{Fig.\XS}

\def\etal{\textit{et al.}\XS}
\def\etc{\textit{etc.}\xspace}
\def\ie{\textit{i.e.,}\xspace}

\def\figuredraft{false}
\def\rev#1{{#1}}

\journal{}

\begin{document}
\begin{frontmatter}
\title{3DTeethSeg'22: 3D Teeth Scan Segmentation and Labeling Challenge}
\author[crns,udini]{Achraf Ben-Hamadou\corref{correspondingauthor}}
\ead{achraf.benhamadou@crns.rnrt.tn}
\author[udini]{Oussama Smaoui}
\author[crns, udini]{Ahmed Rekik}
\author[inria]{Sergi Pujades}
\author[inria]{Edmond Boyer}

\address[crns]{Centre de Recherche en Num\'{e}rique de Sfax,  Laboratory of Signals, Systems, Artificial Intelligence and Networks, Technopôle de Sfax, 3021 Sfax, Tunisia}
\address[udini]{Udini, 37 BD Aristide Briand, 13100 Aix-En-Provence, France}
\address[inria]{Inria, Univ. Grenoble Alpes, CNRS, Grenoble INP, LJK, France}
\author[cgip]{Hoyeon Lim}
\author[cgip]{Minchang Kim}
\author[cgip]{Minkyung Lee}
\author[cgip_ssu]{Minyoung Chung}
\author[cgip]{Yeong-Gil Shin}
\address[cgip]{Department of Computer Science and Engineering, Seoul National University, Republic of Korea }
\address[cgip_ssu]{School of Software, Soongsil University, Republic of Korea}

\author[uncc]{Mathieu Leclercq}
\author[umaa]{Lucia Cevidanes}
\author[uncc]{Juan Carlos Prieto}
\address[umaa]{University of Michigan at Ann Arbor}
\address[uncc]{University of North Carolina at Chapel Hill, USA}
\author[igip]{Shaojie Zhuang}
\author[igip]{Guangshun Wei}
\author[igip_sh]{Zhiming Cui}
\author[igip]{Yuanfeng Zhou}
\address[igip]{Shandong University, China}
\address[igip_sh]{ShanghaiTech University, China}
\author[teethseg]{Tudor Dascalu}
\author[teethseg]{Bulat Ibragimov}
\address[teethseg]{Department of Computer Science, University of Copenhagen, Denmark}
\author[osteam]{Tae-Hoon Yong}
\author[osteam]{Hong-Gi Ahn}
\author[osteam]{Wan Kim}
\author[osteam]{Jae-Hwan Han}
\author[osteam]{Byungsun Choi}
\address[osteam]{Osstem Implant Co., Ltd., Seoul, Korea}
\author[domf,docs,bihc]{Niels van Nistelrooij}
\author[domf,doai]{Steven Kempers}
\author[domf,doai]{Shankeeth Vinayahalingam}
\address[domf]{Department of Oral and Maxillofacial Surgery, Radboud University Nijmegen Medical Centre, P.O. Box 9101, 6500 HB, Nijmegen, the Netherlands}
\address[docs]{Department of Computing Science, Radboud University, Nijmegen, the Netherlands}
\address[doai]{Department of Artificial Intelligence, Radboud University, Nijmegen, the Netherlands}
\address[bihc]{Berlin Institute of Health at Charité – Universitätsmedizin Berlin, Charitéplatz 1, 10117 Berlin, Germany}

\author[udini]{Julien Strippoli}
\author[udini]{Aur\'elien Thollot}
\author[udini]{Hugo Setbon}
\author[udini]{Cyril Trosset}
\author[udini]{Edouard Ladroit}
\begin{abstract}
Teeth localization, segmentation, and labeling from intra-oral 3D scans are essential tasks in modern dentistry to enhance dental diagnostics, treatment planning, and population-based studies on oral health. However, developing automated algorithms for teeth analysis presents significant challenges due to variations in dental anatomy, imaging protocols, and limited availability of publicly accessible data. To address these challenges, the 3DTeethSeg'22 challenge was organized in conjunction with the International Conference on Medical Image Computing and Computer Assisted Intervention (MICCAI) in 2022, with a call for algorithms tackling teeth localization, segmentation, and labeling from intraoral 3D scans. A dataset comprising a total of 1800 scans from 900 patients was prepared, and each tooth was individually annotated by a human-machine hybrid algorithm. A total of 6 algorithms were evaluated on this dataset. In this study, we present the evaluation results of the 3DTeethSeg'22 challenge. The 3DTeethSeg'22 challenge code can be accessed at: \href{https://github.com/abenhamadou/3DTeethSeg22_challenge}{https://github.com/abenhamadou/3DTeethSeg22\_challenge}.
\end{abstract}

\begin{keyword}
Teeth localization, \sep 3D Teeth segmentation, 3D segmentation, 3D object detection, 3D intraoral scans dentistry
\end{keyword}
\end{frontmatter}

\section{Introduction}\label{sec:intro}
Computer-aided design (CAD) tools have become increasingly popular in modern dentistry for highly accurate
treatment planning. In particular, in orthodontic CAD systems, advanced intraoral scanners (IOSs) are now widely used as they provide precise digital surface models of the dentition. Such models can dramatically help dentists simulate teeth extraction, move, deletion, and rearrangement and ease therefore the prediction of treatment outcomes. Hence, digital teeth models have the potential to release dentists from otherwise tedious and time consuming tasks.
Although IOSs are becoming widespread in clinical dental practice, there are only few contributions on teeth
segmentation/labeling available in the literature \cite{lian2019meshsnet,Xuetal2018,sun2020automatic} and no publicly available database. A fundamental issue
that appears with IOS data is the ability to reliably segment and identify teeth in scanned observations.
Teeth segmentation and labeling is difficult as a result of the inherent similarities between teeth shapes as well as
their ambiguous positions on jaws.
In addition, it faces several challenges:
\begin{enumerate}
\item  The teeth position and shape variation across subjects.
\item  The presence of abnormalities in dentition. For example, teeth crowding which results in teeth misalignment
and thus non-explicit boundaries between neighboring teeth. Moreover, lacking teeth and holes are commonly
seen among people.
\item  Damaged teeth.
\item  The presence of braces, and other dental equipment.
The challenge we propose will particularly focus on point 1, \ie the teeth position and shape variation across
subjects. With the extension of available data in the mid and long term, the other points will also be addressed in
further editions of the challenge.

\end{enumerate}
\subsection{Terminology}
In this section, we will explore three essential terms used in the analysis of intraoral 3D scans: localization, segmentation, and labeling. Localization refers to the precise identification and positioning of a tooth, including the calculation of its 3D centroid. Segmentation, on the other hand, involves identifying the vertices that pertain to a detected tooth, allowing for the demarcation of its boundaries. Labeling involves assigning a specific class to a detected and segmented tooth. In this work, we adhere to the FDI teeth numbering system.
\subsection{Prior work}
The majority of relevant works on the topic fall into two categories: handcrafted feature-based approaches and learning-based approaches.

\subsection{Handcrafted features-based approaches}
Former approaches were mainly grounded on the extraction of handcrafted geometric features to segment 3D dental scans. These approaches are broadly classified into three types: surface curvature-based methods, contour line-based methods, and harmonic field-based methods. Surface curvature is highly informative in IOSs for characterizing tooth surfaces and locating tooth/gum borders. This feature is used in \citep{zhao2006interactive} to propose a semiautomatic method for teeth segmentation based on curvature thresholding, in which gum segregation is followed by 3D teeth boundary curve identification to ensure the segmentation process. Later, Yuan \etal developed an integrated single-tooth modeling scheme for region extraction and teeth separation based on surface minimum curvature calculation \citep{old_ios_seg}. Wu \etal \citep{wu2014tooth} proposed a Morphological skeleton-based method for teeth segmentation from IOSs by separating teeth using area growing operations. In the same vein, \citep{kronfeld2010snake} suggested a system for detecting the boundaries between teeth and gingiva based on active contour models. For the contour-line method, tooth boundary landmarks are manually selected by users on dental 3D scans. In \citep{sinth,yaqi2010computer}, for instance,  the selected tooth boundary is used to calculate the contour lines from their geodesic information and generate the desired final tooth boundaries. 
We can also mention the harmonic-field method which is more user-friendly for teeth segmentation than the previous approaches. In comparison to previous techniques, this method requires less user interaction by allowing them to select a limited number of surface points prior to the segmentation process \citep{zou2015interactive,liao2015automatic}.

The approaches described above fall short when it comes to robust and fully automated segmentation of dental 3D scans. Setting the optimal threshold value for surface curvature-based methods is not straightforward. Indeed, these methods are still sensitive to noise, and selecting the wrong threshold can systematically affect the segmentation accuracy, resulting in an over-or under-segmentation problem. Furthermore, the manual threshold selection will always make curvature-based methods far from being applied in a fully automatic mode. Also, contour-line approaches are time-consuming, tough to use, and closely rely on human interaction. Finally, the harmonic field techniques involve sophisticated and heavy pre-processing steps.

\subsection{Learning based approaches}

Teeth segmentation techniques have recently shifted away from hand-crafted features and toward learned features thanks to deep learning techniques. Indeed, it is nowadays crystal clear that data-driven feature extraction, using CNNs for example, outperforms handcrafted features for many computer vision tasks, such as object detection \citep{ren2015faster}, image classification \citep{wang2016cnn}, \etc and 3D teeth segmentation and labeling is no exception. Depending on the input data, features learning methods methods can be divided into two main approaches: 2D image segmentation and 3D mesh segmentation. CNNs have been used in numerous studies to extract relevant features from 2D images. Particularly, Cui \etal \citep{cui2019toothnet} introduced a two-stage deep supervised neural network architecture for automatic tooth instance segmentation and identification from Cone-Beam Computed Tomography (CBCT) images. A set of edge maps were first extracted from the CBCT slices with an autoencoder CNN and then fed to a Mask R-CNN network \citep{8237584} for tooth segmentation and recognition.
Another study fine-tuned  a pre-trained AlexNet network on CBCT dental slices for automatic teeth classification \citep{miki2017classification}. 
A symmetric fully convolutional residual neural network was suggested by \citep{rao2020symmetric} to generate a segmentation probability map for teeth in CBCT images. Following that, the dense conditional random field technique and the deep bottleneck architecture were used for teeth boundary smoothing and segmentation enhancement, respectively. Zhang \etal \citep{zhang2020automatic} isomorphically mapped 3D dental scans into 2D harmonic parameter space to generate 2D images that will be fed into a CNN-based on the U-Net architecture for tooth image segmentation.

Taking advantage of recent advances in deep learning techniques and hardware computing capability, researchers started to employ deep learning-based methods directly on 3D dental meshes. Sun \etal used FeaStNet \citep{vermaFeaStNetFeatureSteeredGraph2018} a graph CNN-based architecture for automated tooth segmentation and labeling from 3D dental scans \citep{sun2020tooth}. They then extended the previous architecture and proposed an end-to-end graph convolutional network-based model for tooth segmentation and dense correspondence of 3D dental scans, in which a geodesic map and a probability matrix were used to improve the segmentation performance \citep{sun2020automatic}. Xu \etal \citep{Xuetal2018} introduced a multi-stage framework based on deep CNN architecture for 3D dental mesh segmentation where teeth-gingiva and inter-teeth labeling processes were achieved by training two independent CNNs. Similarly, Zanjani \etal proposed an end-to-end deep learning system based on the PointNet \citep{PointNet2017} network architecture for semantic segmentation of individual teeth and gingiva from point clouds representation, as well as a secondary neural network as a discriminator in an adversarial learning setting for teeth labeling refinement \citep{zanjani2019deep}. A broader perspective has been adopted by Lian \etal who modified the original version of the PointNet architecture to incorporate a set of graph-constrained learning modules in order to extract multi-scale local contextual features for teeth segmentation and labeling on 3D Intra-Oral-Scans \citep{lian2020deep}. Differently to \citep{lian2020deep,zanjani2019deep}, authors in  \citep{tian2019automatic} added a preprocessing step to encode the input 3D scans using a sparse voxel octree partitioning before separately feeding a three-level hierarchical CNNs learning for the segmentation process and another two-level hierarchical CNNs for the teeth recognition. A different approach was recently proposed in  \citep{cui2020tsegnet} where a pipeline is divided into two key components: a first CNN dedicated to 3D centroids prediction as a teeth localization, followed by a second CNN applied separately on each pre-localized tooth crop for joint tooth/gum segmentation and tooth type recognition. In the same vein, in \citep{Zanjanietal2019} a region proposal network (RPN) based on a Monte Carlo approach was proposed as the first step of teeth localization. This RPN is followed by a Mask R-CNN-like architecture for instance wise teeth segmentation. Finally, as a post-processing procedure, a look-up table on the teeth centroids assigns the labels to the detected teeth.
Another deep neural network architecture was suggested in \citep{ma2020srf} for pre-detected teeth classification on 3D scanned point clouds based on adjacency similarity and relative position features vectors. It attempts to explicitly model the spatial relationship between adjacent teeth for recognition.

Zhao \etal \citep{zhao_3d_2021} proposed an end-to-end network that adopts a series of graph attentional convolution layers and a global structure branch to extract fine-grained local geometric features and global features from raw mesh data. Then, these features are fused to learn the segmentation and labeling tasks. Zhao \etal \citep{zhao_two-stream_2022} suggested a two-stream graph convolutional network (TSGCN) where the first stream captured coarse structures of teeth from the mesh 3D coordinates, while the second stream extracted distinctive structural details from its normal vectors.
Since current learning-based methods mainly rely on expensive point-wise annotations, Qiu \etal \citep{qiu2022darch} introduced the Dental Arch (DArch) method for 3D tooth instance segmentation using weak low-cost annotated data (labeling all tooth centroids and only a few teeth for each dental scan).  The DArch consists of two stages: tooth centroid detection and tooth instance segmentation where the dental arch is initially generated by Bezier curve regression, and then refined using a graph-based convolutional network (GCN).

\section{Materials and challenge setup}\label{sec:materials}
%
%
\subsection{Data acquisition and annotation}
    \subsubsection{Data acquisition}
    In compliance with the European General Data Protection Regulation (GDPR) agreement, we obtained 3D intra-oral scans for 900 patients acquired by orthodontists/dental surgeons with more than 5 years of professional experience from partner dental clinics located mainly in France and Belgium. All data is completely anonymized, and the identity of the patients cannot be revealed. Two 3D scans are acquired for each patient, covering the upper and lower jaws separately. The following IOSs were used for scan acquisition: the Primescan from Dentsply, the Trios3 from 3Shape, and the iTero Element 2 Plus. These scanners are representative and generate 3D scans with an accuracy between 10 and 90 micrometers and a point resolution between 30 and 80 pts/mm2. No additional equipment other than the IOS itself was used  during the acquisitions. All acquired clinical data are collected for patients requiring either orthodontic ($50\%$) or prosthetic treatment ($50\%$). The provided dataset follows a real-world patient age distribution: $50\%$ male $50\%$ female, about $70\%$ under 16 years-old, about $27\%$ between 16-59 years-old, about $3\%$ over 60 years old.
    
    \subsubsection{Data annotation and processing}
\begin{figure}[!h]
        \centering
        \includegraphics[draft=\figuredraft,width=0.6\textwidth]{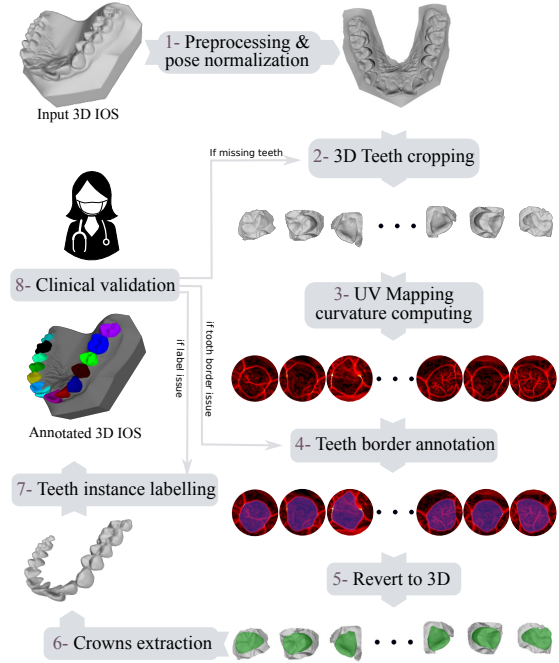}
        \caption{Illustration of our annotation process. An input 3D IOS is annotated following eight steps, beginning with preprocessing and pose normalization and ending with clinical validation. The clinical validator can return the annotation to steps 2, 4, or 7, depending on the raised issue, which respectively corresponds to missing teeth, teeth border issues, or incorrect teeth instance labeling.}   
        \label{fig:annotation_process}
    \end{figure}

    The data annotation, \ie teeth segmentation and labeling, was performed in collaboration with clinical evaluators with more than 10 years of expertise in orthodontistry, dental surgery, and endodontics. The detailed process is depicted in \figureabvr \ref{fig:annotation_process}.
    It consists of eight steps. First, the 3D scans are preprocessed (steps 1 and 2 in \figureabvr \ref{fig:annotation_process}) by removing all degenerated and redundant mesh faces, as well as duplicated and irrelevant vertices. The dental mesh coordinates are then automatically centered and aligned with the occlusal plane by principle component analysis. This improves teeth visibility while also normalizing the 3D pose of all the input 3D IOSs. Then, we used a custom tool to manually crop each tooth from the 3D scans with a tight sphere that includes the detected tooth as well as its surroundings (\ie neighboring teeth and gingiva). We decided to perform UV mapping in step 3 to flatten the cropped 3D meshes and show the maximum 3D curvature to make the annotation of tooth boundaries easier. This transformation is ensured by harmonic parameterization, a fixed boundary parameterization algorithm that calculates the 2D coordinates of the flattened cropped tooth as two harmonic functions \cite{eck1995multiresolution}. These vertices are then mapped to a circle, and the 2D coordinates of the remaining vertices are calculated using two harmonic functions and the circle boundary constraints. The benefits are twofold: the annotator can annotate the 2D polygons delimiting the tooth without changing the 3D point of view, and the 3D curvature overlay is informative on the boundaries of teeth.
    
\begin{figure}[!h]
        \centering
        \includegraphics[draft=\figuredraft,width=0.6\textwidth]{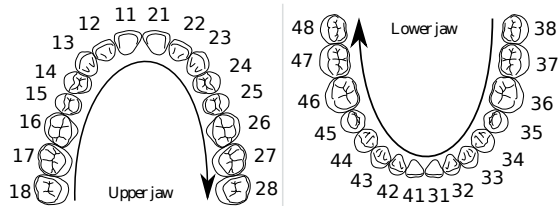}
        \caption[]{FDI World Dental Federation notation\footnotemark.}
        \label{fig:fdi_teeth_labels}
    \end{figure}

    After the manual annotation of the UV maps (step 4 of \figureabvr \ref{fig:annotation_process}), we back-propagate the tooth boundaries to the 3D crops in step 5. At this point each separate tooth candidate has been manually segmented, however they have been represented in the same 3D coordinate system. The aim of the next step is to gather all the teeth crowns and prepare them for manual labeling as shown in step 7 of \figureabvr \ref{fig:annotation_process}. We followed the FDI World Dental Federation numbering system for teeth labeling depicted in \figureabvr \ref{fig:fdi_teeth_labels}.
    \footnotetext{FDI World Dental Federation numbering system, Wikipedia, last modified May 20, 2022.\\ https://en.wikipedia.org/wiki/FDI$\_$World$\_$Dental$\_$Federation$\_$notation}
    The final step 8 of the annotation process consists of the visual inspection and validation of the produced annotated 3D IOS. This step is carried out by our clinical partners, who are experienced orthodontists, dental surgeons, and endodontists. Their inspection targeted the identification of annotation issues, such as a missing tooth annotation (return to step 2), inaccurate  tooth boundary annotation (return to step 4), or incorrect tooth labels (return to step 7). This validation/correction cycle was repeated until the entire dataset was correctly annotated and clinically validated.
    \section{Data records}\label{sec:ResultingDataset}
\begin{figure}[t]
        \centering
        \includegraphics[draft=\figuredraft,width=1\textwidth]{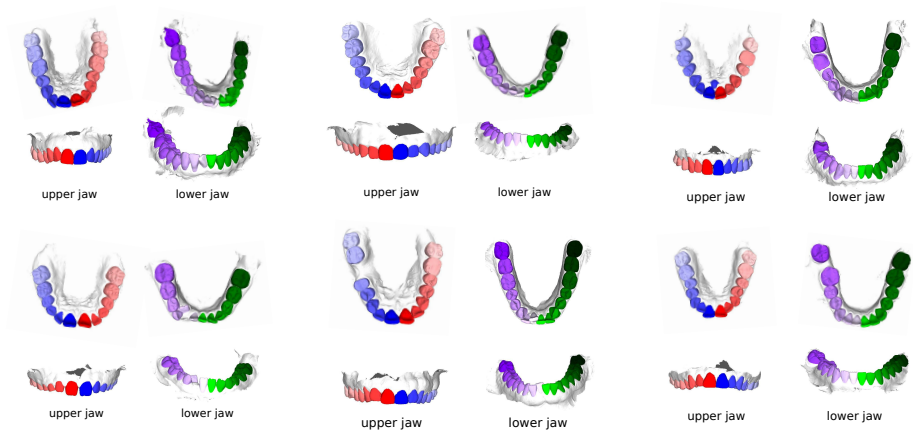} 
        \caption{Frontal and occlusal rendering of annotated jaws for 6 randomly selected patients.}\label{fig:examples}
    \end{figure}

    A total of 1800 3D intra-oral scans have been collected for 900 patients covering their upper and lower jaws separately. \figureabvr \ref{fig:examples} shows some examples.
    The data are hosted at the Figshare platform \footnote{The link is provided in \href{https://github.com/abenhamadou/3DTeethSeg22_challenge}{https://github.com/abenhamadou/3DTeethSeg22\_challenge}}. For the purpose of the segmentation and labeling challenge held at MICCAI 2022, the 3D scans data (obj format) are separated into training (1200 scans, 16004 teeth)  and test  data (600 scans, 7995 teeth). Patients are anonymized by their universally unique identifier (uuid). The ground truth tooth labels and tooth instances for each vertex in the obj files are provided in JavaScript Object  Notation (JSON) format. A JSON file example is shown below:
    \begin{lstlisting}[language=json,firstnumber=1]
    {
        "id_patient": "YNKZHRPO", 
        "jaw": "upper",
        "labels":    [0, 0, 44, 33, 34, 0, 0, 45, 0, ..., 41, 0, 0, 37, 0, 34, 45, 0, 31, 36], 
        "instances": [0, 0, 10,  2, 12, 0, 0,  9, 0, ..., 10, 0, 0,  8, 0,  1,  9, 0,  1,  8]
    }
    \end{lstlisting}
    The length of the tables "labels" and "instances" is the same as the total number of vertices in the corresponding 3D scan. The label and instance "0" are reserved by default for gingiva. And, other than "0", the unique numbers in table "instances" indicate the number of teeth in the 3D scan.
     
    \subsection{Challenge setup}

The 3DTeethSeg'22 challenge was organized as a Satellite Event at MICCAI 2022, with a specific focus on algorithms addressing teeth detection, segmentation, and labeling using intra-oral 3D scans. The challenge comprised three distinct phases: one training phase and two testing phases. During the training stage, participants were provided access to the scans along with their corresponding annotations. This allowed them to design and train their algorithms using the provided data. The first testing phase involved a preliminary evaluation, where participants could submit their algorithms and assess their performance on a limited dataset of 10 scans. In the final testing phase, participants were not granted access to the test scans directly. Instead, they were required to submit their code within a docker container. The dockers were then evaluated on hidden test data, preventing any retraining on the test data or overfitting through fine-tuning. All the training and testing data, along with their annotations, are now publicly available. Additionally, we have open-sourced the data processing and evaluation scripts to facilitate further research and development. All the materials related to the 3DTeethSeg'22 challenge can be accessed at the following link: \href{https://github.com/abenhamadou/3DTeethSeg22_challenge}{https://github.com/abenhamadou/3DTeethSeg22\_challenge}

    \subsection{Evaluation metrics \label{ses:evaluation_metrics}}
    
            \subsubsection{Teeth localization accuracy}
            -Teeth localization accuracy (TLA): mean of normalized Euclidean distance between ground truth (GT) teeth centroids and the closest localized teeth centroid. Each computed Euclidean distance is normalized by the size of the corresponding GT tooth. In case of no centroid (e.g. algorithm crashes or missing output for a given scan) a nominal penalty of 5 per GT tooth will be given. This corresponds to a distance 5 times the actual GT tooth size. As the number of teeth per patient may be variable, here the mean is computed over all gathered GT Teeth in the two testing sets. 
            
            \subsubsection{Teeth segmentation accuracy}
            - Teeth segmentation accuracy (TSA): is computed as the average F1-score over all instances of teeth point clouds. The F1-score of each tooth instance is measured as: $$F1=2*\frac{precision \times recall}{precision+recall}$$
            
            \subsubsection{Teeth identification rate}
            - Teeth identification rate (TIR): is computed as the percentage of true identification cases relative to all GT teeth in the two testing sets. A true identification is considered when for a given GT Tooth, the closest detected tooth centroid: is localized at a distance under half of the GT tooth size, and is attributed the same label as the GT tooth 

\section{Methods}\label{sec:methods}
Over five hundred requests for data download and registration have been received for the 3DTeethSeg'22 challenge. Forty-four teams registered and participated to the preliminary phase, and only ten teams uploaded their submissions to the leader board for the final phase of the challenge. Table \ref{tab:methods_summary} provides a brief synopsis of the selected six top ranked methods that will be presented in more details below in this section.

\FloatBarrier
\begin{sidewaystable}[t]
    \caption{Brief summary of the participating methods in 3DTeethSeg'22 challenge, \rev{ordered alphabetically according to referring author.}}
    \centering
    \begin{tabular}{lp{10cm}}
    \toprule
    Team/Ref. Authors  & Method features \\
    \fcolorbox{white}{pink}{\rule{0pt}{5pt}\rule{5pt}{0pt}}\quad CGIP team - Hoyeon Lim \etal & Multi-stage, tooth instance module labels for each vertex of the dental mesh. Point Grouping Module based on Point Transformer backbone network to predict tooth semantic labels and offsets. Teeth cropping module infers the tooth-gingiva mask to refine the tooth instance labels. \\
    \fcolorbox{white}{teal}{\rule{0pt}{5pt}\rule{5pt}{0pt}}\quad FiboSeg team - Mathieu Leclercq \etal &  A Residual U-Net model trained on rendered 2D views of dental models, where normal vectors are encoded as RGB components. A majority voting scheme assigns labels to each face in the dental model, followed by post-processing for for \textit{island} removal and boundary smoothing.    \\
    \fcolorbox{white}{olive}{\rule{0pt}{5pt}\rule{5pt}{0pt}}\quad IGIP team - Shaojie Zhuang \etal & Multi-stage, teeth Tooth-gingiva separation, Tooth centroid prediction, individual teeth segmentation from patches cropped around the predicted centroids, teeth classification combining shape and position features, post-processing stage using dental arch curve to correct classification errors.\\
    \fcolorbox{white}{magenta}{\rule{0pt}{5pt}\rule{5pt}{0pt}}\quad TeethSeg team - Tudor Dascalu \etal & Multi-stage, coarse segmentation phase using a 3D U-net model and a fine segmentation stage that utilized local geometric features to enhance accuracy around tooth-tooth and tooth-gum edges. The Random Walker algorithm was employed to label the vertices of the dental cast, taking into account the convexity of the neighborhood and minimizing movement over edge regions. \\
    \fcolorbox{white}{lime}{\rule{0pt}{5pt}\rule{5pt}{0pt}}\quad  OS team - Tae-Hoon Yong \etal & Two stage approach, high-resolution network architecture with sub-networks operating in parallel to predict teeth centroids from 2D images and  rendered for the top view of the input 3D mesh.  Individual tooth cropping and segmentation based on the results of the first stage, and the segmented tooth regions were aggregated and mapped back to the original mesh. The final step involved upsampling the decimated models to the original scan model using the K-Nearest Neighbor algorithm.\\
    \fcolorbox{white}{purple}{\rule{0pt}{5pt}\rule{5pt}{0pt}}\quad Champers team - Niels van Nistelrooij \etal & Multi-stage, encoder-decoder network based on the Stratified Transformer architecture for tooth centroid prediction. The tooth segmentation stage employed a cascade of two Stratified Transformers to segment individual teeth. Post-processing steps were applied to merge proposals, interpolate segmentations, and assign labels to the source point cloud. 
    \end{tabular}
    \label{tab:methods_summary}
\end{sidewaystable}
\FloatBarrier

%
\subsection{CGIP Team - Hoyeon Lim \etal}
\begin{figure*}[!t]
\centering
\includegraphics[width=\textwidth]{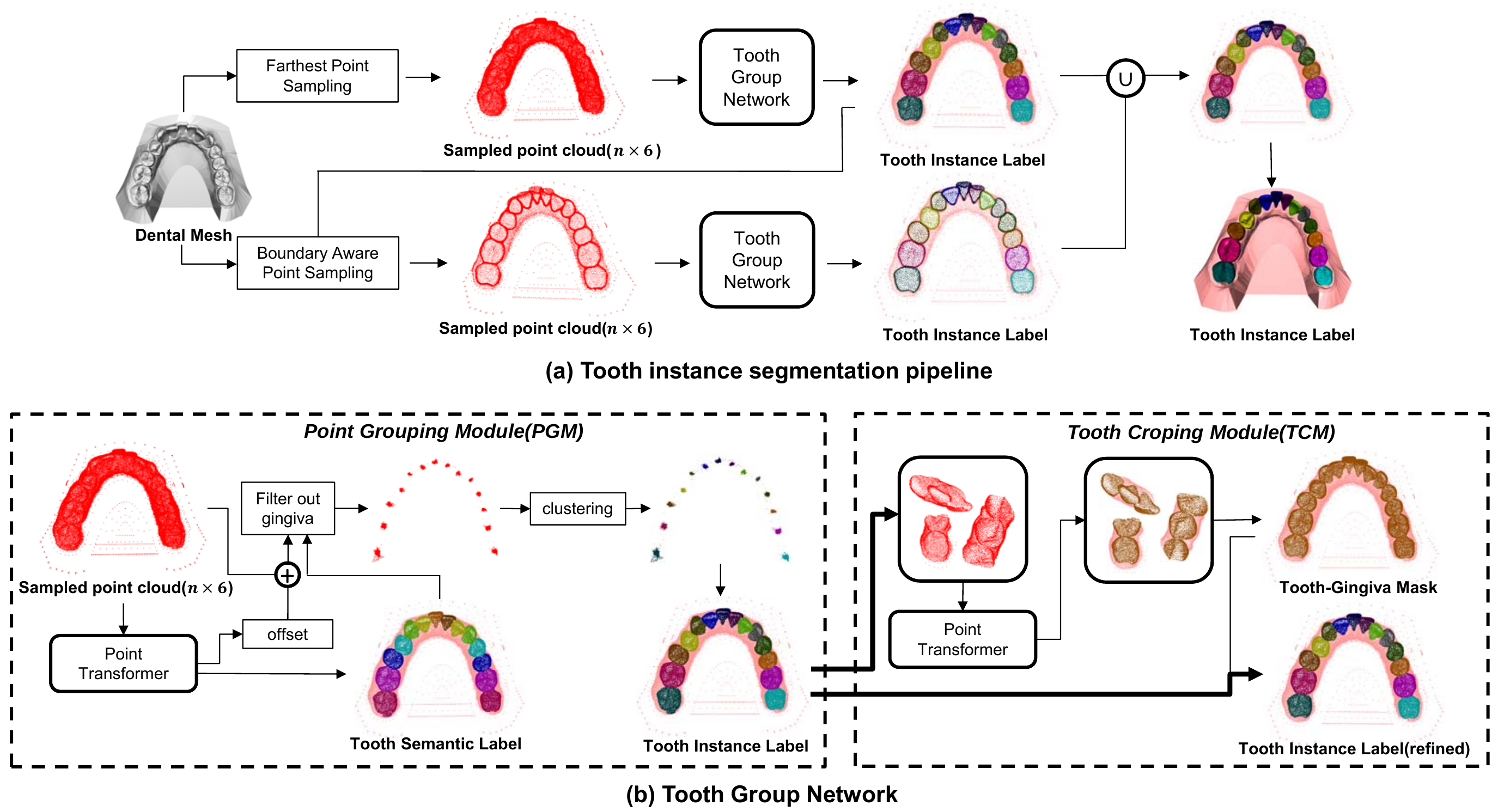}
\caption{Proposed approach by CGIP team. (a) Tooth instance segmentation pipeline predicts tooth instance labels for each vertex of the dental mesh. (b) Point Grouping Module takes sampled point cloud and outputs tooth semantic labels and tooth instance labels. These tooth instance labels are then refined by Tooth Cropping Module. The color of the tooth instance label represents each individual tooth instance, while the color of the tooth semantic label indicates the tooth class. For example, the green point in the tooth semantic label corresponds to the canine tooth.}
\label{fig:cgip_team_framework}
\end{figure*}

As shown in \figureabvr \ref{fig:cgip_team_framework}, the proposed method consists of two main modules. First, a tooth instance segmentation pipeline predicts tooth instance labels for each vertex of the dental mesh. Then, the Point Grouping Module takes a sampled point cloud and outputs tooth semantic labels and tooth instance labels.

\subsubsection{Tooth instance segmentation pipeline}
As shown in Figure \ref{fig:cgip_team_framework}(a), the proposed tooth instance segmentation pipeline takes the dental mesh and outputs tooth instance labels for each vertex of the dental mesh. The Tooth Group Network accepts features of sampled points and generates tooth instance labels. Two sampling methods to obtain the sampled points.

To obtain the final instance segmentation result, it is necessary to predict all tooth instance labels for each vertex of the dental mesh. The tooth instance label of a non-sampled point is determined by assigning it the label of the nearest neighbor point. Due to the nature of the 3D scanner, the sampling rate of the dental mesh is high near the boundary. Therefore, points near the boundary may be associated with multiple labels, which prevents obtaining fine-grained tooth instance labels. To address this issue, a solution called the Boundary Aware Point Sampling method has been proposed. This method aims to increase the number of sampled points near the boundary. Initially, \(n\) points are sampled by utilizing the Farthest Point Sampling technique on the vertices of the dental mesh. The Tooth Group Network takes these sampled points as input and generates tooth instance labels for them. By examining the predicted tooth instance labels, points located in close proximity to the boundary can be identified. Subsequently, additional points are sampled near the boundary using the Boundary Aware Point Sampling approach. Another instance of the Tooth Group Network is then employed to generate tooth instance labels for these newly sampled points situated near the boundary.

To derive the final tooth instance segmentation result, tooth instance labels are aggregated from both Farthest Point Sampling and Boundary Aware Point Sampling. The class of each tooth instance can be determined by conducting a majority vote among the tooth semantic labels within the corresponding tooth instance. The tooth semantic labels are obtained by inputting the points sampled through Farthest Point Sampling into the Tooth Group Network (see \figureabvr 1(b)).
\subsubsection{Tooth Group Network}
As depicted in \figureabvr \ref{fig:cgip_team_framework}(b), The Tooth Group Network is composed of Point Grouping Module(PGM) and Tooth Cropping Module(TCM). The backbone network of PGM and TCM is Point Transformer\citep{Zhaoetal2021}. The difference is that PGM has a regression head for offset prediction and a classification head for tooth semantic label prediction while TCM has just one head to infer tooth-gingiva mask. 

PGM has a similar process to Point Group\citep{Jiangetal2020}. The backbone network of PGM takes sampled point cloud that contains coordinates and normals of \(n\) points. It predicts tooth semantic label and offset. Then, the shifted point cloud is obtained by moving each point to its center point according to offsets. Points that are predicted as gingiva are filtered out from this point cloud. The tooth instance labels of each point are obtained by clustering the shifted point cloud because points closer to each other in the shifted point cloud are likely to belong to the same instance. The DBSCAN\citep{Esteretal1996} is used to group 3d points. The clustering-based tooth instance labeling process is robust because each tooth instance has inherently a compact cylinder shape that is easy to group.

In TCM, the sampled point cloud is cropped around the center points of predicted tooth instances and feed them into the backbone network of TCM. As a result, the tooth-gingiva mask is generated as output.  This mask is used to refine tooth instance labels of PGM. The tooth instance label of the point is changed to gingiva if it is predicted as a tooth in PGM but predicted as gingiva in TCM. If the tooth instance label of the point is predicted as gingiva in PGM but predicted as tooth in TCM, the tooth instance label of the point is determined by the label of the nearest neighbor point.

To prevent a decrease in tooth instance segmentation accuracy near the boundary where the tooth segmentation label changes, the contrastive boundary learning framework\citep{Tangetal2022} is adopted. This framework makes two points near the boundary have similar features if they have the same label. Conversely, if they have different labels, the backbone network learns to output contrastive features for the two points. The regression head for offset and classification head for tooth semantic label can take advantage of these distinct features when the heads predicts tooth instance labels near the boundary.
%
%

\subsection{FiboSeg Team - Mathieu Leclercq \etal}
\begin{figure*}[!t]
\centering
\includegraphics[width=0.95\textwidth]{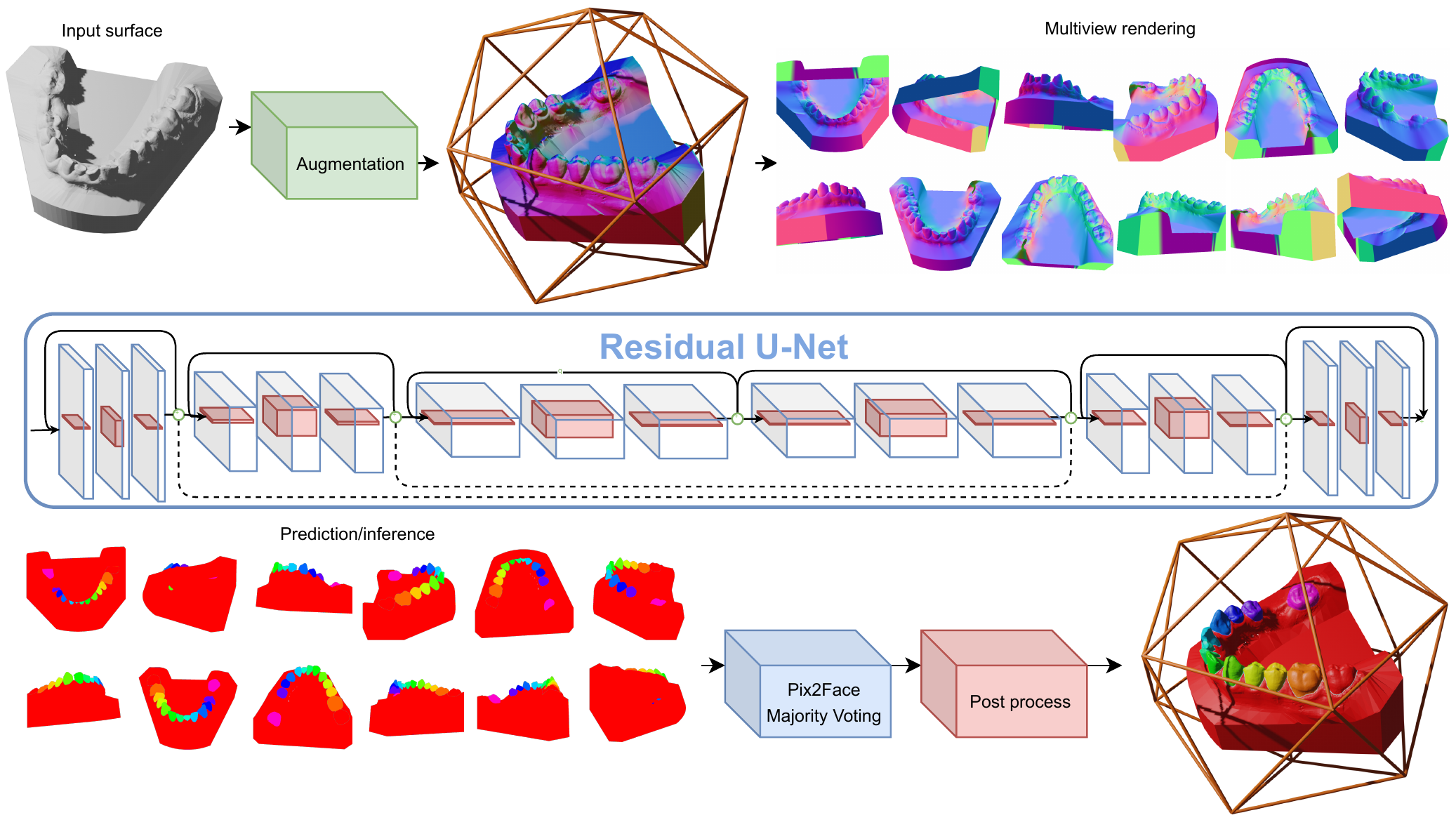}
\caption{Proposed approach by FiboSeg team. During training, the dental models are randomly rotated and random teeth are removed. The views are rendered from different viewpoints and colored using normal vectors encoded in the RGB components as well as a depth map used as a fourth component. The ground truth target images are also created during the rendering step. The images are fed to a Residual U-Net and the output and target images are used to compute the DiceCELoss. At inference time, a majority voting scheme is employed to assign labels to each face in the dental model. Subsequently, an island removal approach is applied to eliminate isolated regions, and the output boundaries of the segmented teeth are smoothed for a more refined result.}
\label{fig:fiboseg_team_framework}
\end{figure*}

This work presents a deep learning-based method for 3d dental model segmentation. It consists of acquiring 2D views and extracting features from the surface such as the normal vectors. The rendered images are analyzed with a 2D convolutional neural network, such as a U-NET.

\subsubsection{Rendering the 2D views}
The Pytorch3D framework is used for the rendering of the 3D intraoral surface model from different viewpoints. 
The views are fed to a residual U-Net\cite{hatamizadeh2022unetr} in an end-to-end training procedure. The rendering engine provides a map that relates pixels in the images to faces in the mesh and allows rapid extraction of point data (normals, curvatures, labels, etc.) 
as well as setting information back into the mesh after inference. 
In order to get different views, random rotations are applied to the camera, so that it moves on the surface of a unit sphere. 
For each snapshot, we generate two images. The first one contains the surface normals encoded in the RGB components + a depth map. The second one contains the ground truth label maps that are used as targets in the segmentation task. 
We set the resolution of the rendered images to 320px. We use ambient lights so that the rendered images don't have any specular components.

\subsubsection{Training of the network}
We augment the data by applying random rotations and randomly removing dental crowns (excluding wisdom teeth).  
The residual U-Net model is instantiated using the MONAI\footnote{monai.io} library with 4 input channels, 34 output channels, 
and encoder/decoder blocks with 2 residual units and channels 16, 32, 64, 128, 256 with stride 2. 

We use the DiceCELoss, which computes the Dice Loss as well as Cross-Entropy Loss and returns the weighted sum of these two. 

\begin{equation}
    DiceCELoss = w_0\frac{2\sum_{c=1}^N p_c y_c}{\sum_{c=1}^N p_c^2 + \sum_{c=1}^N p_c^2 y_c^2}  - w_1\sum_{c=1}^N y_c\log(p_c)
\end{equation}

The learning rate is set to $1e^-4$ using the AdamW optimizer. One important thing to note is that there is no previous pre-processing to the mesh, \textit{i.e.}, sub-sampling of points/faces, or any classification task to identify upper or lower jaws. 
The training learns to identify 34 different labels corresponding to the upper and lower crowns. 
We use one-hot encoding for the 34 different classes: 32 different crowns, in addition to the gum and the background.

\subsubsection{Prediction}
The prediction has three major steps: 1. Render 2D views from the 3D object; 2. Run inference on the 2D views. 3. map the 
information back into the 3D mesh.
The Pytorch3d rasterizer returns a mapping that keeps track of the nearest face at each pixel, 
after we run inference on the 2D views, we use a weighted majority voting scheme to put information back into the 3D mesh
as a single face may be rendered by 2 separate views.
Faces that are hit by zero pixels are given the value $-1$.  

\subsubsection{Post-Processing}
In the event that some faces of the surface are not assigned to any label 
at the end of the prediction, we apply an `island removal' approach, that assigns the closest-connected label.
Finally, we apply a morphological closing operation to smooth the boundary of the segmented teeth. 
%
%
\subsection{IGIP team - Shaojie Zhuang \etal}
\begin{figure}
    \centering
    \includegraphics[width=\textwidth]{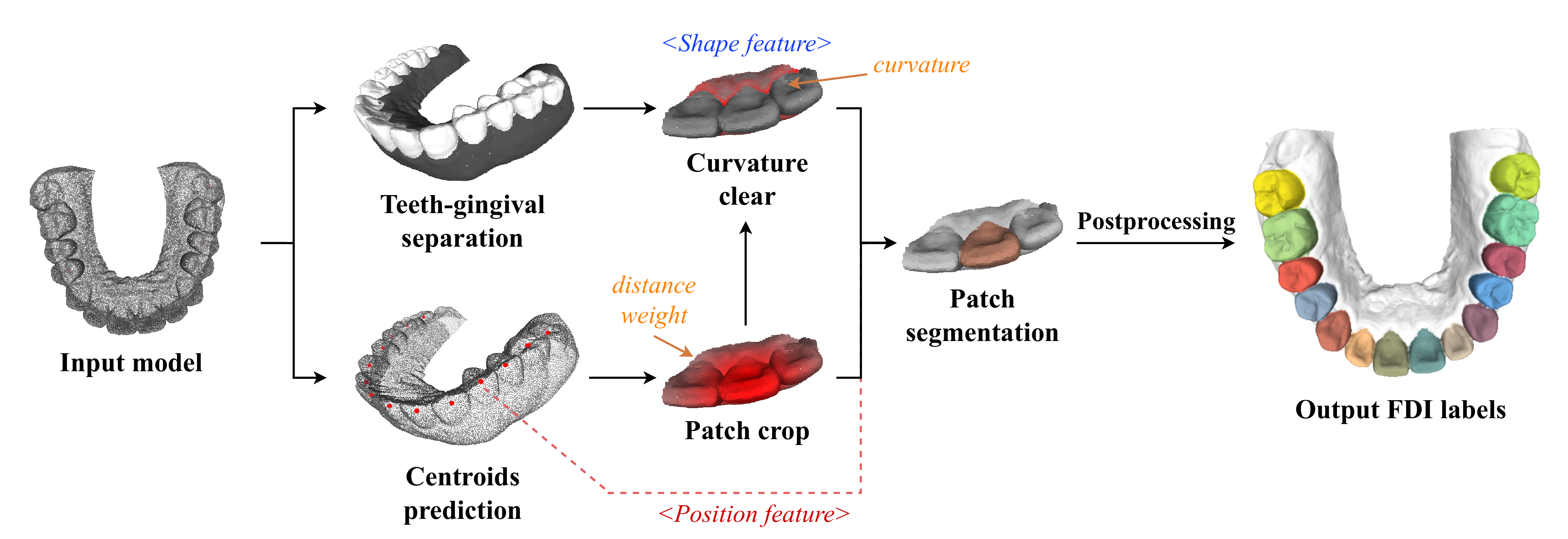}
    \caption{Proposed approach by IGIP team. The centroids are predicted for each tooth, based on which the patches are cropped, and then the curvature on tooth crown is removed. Next, the tooth labels and masks are predicted and mapped back to the original model in patch segmentation stage. Finally, a pos-processing stage is applied to fix error labels.}
    \label{fig:pipeline}
\end{figure}
As shown in \figureabvr~\ref{fig:pipeline}, it is a multi-stage method for accurate teeth segmentation. First, the point cloud is down-sampled to $N = 32768$ points as the input. Then, the input is separated into teeth and gingival. Centroids are predicted on teeth points, around which the patches are cropped, and individual tooth per patch is segmented. Next, the teeth classification is done by combining the local and global features, followed by a post-processing stage to correct potential classification errors.

\noindent\textbf{Teeth-gingival separation.}
For the first step, a binary classification between teeth and gingival is made to remove most non-tooth points in the point cloud, in order to reduce the inference caused by model bases.

\noindent\textbf{Centroids prediction.}
By using PointNet++~\cite{qi2017pointnet++}, the input point cloud $P$ with $N$ points is down-sampled to point cloud $P'$ with $N'$ points. Each point belonging to $P'$ is regressed to obtain an offset vector $\Delta P'$. The predicted centroids are finally obtained by $\hat{C} = P' + \Delta P'$. Similar with~\cite{cui2020tsegnet}, for each predicted centroid $\hat{c}_i \in \hat{C}$, the closest ground truth centroid $c_i$ is its target, and the network uses the following loss functions for supervision:
\begin{equation}
    \begin{split}
        L_{cent} =& \frac{1}{M}\sum_{i = 1}^{M} [L_1^{smooth}(\hat{c}_i - c_i) + \lambda \frac{\left \| \hat{c}_i - c_{i1} \right \|_2 }{\left \| \hat{c}_i - c_{i2} \right \|_2}]\\
        & + L_{CD}(\hat{C}, C),
    \end{split}
\end{equation}
\noindent where the smooth L1 loss supervises the distance between the predicted centroids and the ground truth centroids, $c_{i1}$ and $c_{i2}$ are the ground truth centroids with the minimum and the second minimum distance from $\hat{c}_i$ respectively, used to push a predicted centroid away from other ground truth centroids, $\lambda$ is set to 0.2 after verification, the $L_{CD}$ is the chamfer distance loss. After the prediction, the dense centroids are clustered using density peaks clustering algorithm~\cite{alex2014cfdp} to get the final centroids.

\noindent\textbf{Patch segmentation.}
Patches are cropped around each predicted centroid. The patch size is set to $N/8$, which ensures that at least one full tooth is contained in one patch. A distance weight is attached to each point to mark the tooth, calculated as:
\begin{equation}
    w_{s_i} = e^{-2 \times \left \| s_i - \hat{c}_i \right \|_2},
\end{equation}
where $s_i$ is the $i$-th point in one patch, and $\hat{c}_i$ is the predicted centroid. Besides, the curvature, as a feature, is able to strengthen significance at the teeth-gingival boundary. The curvature at teeth crowns is removed by the binary mask obtained at the first step. The output is supervised by a cross-entropy loss. In addition, in case there exist multiple centroids in a tooth, an overlap detection for patches is made, that if predicted masks of two patches overlap a lot, then they are merged into one patch.

\noindent\textbf{Teeth classification.}
The importance of teeth classification is often ignored in most algorithms. It is a common way to output point-wise labels of a whole model by instance segmentation networks, but such algorithms don't take the features of an individual tooth into account. Some methods like ~\cite{im2022accuracy, lian2019meshsnet}, only classify a tooth into 16, 14 or fewer classes, ignoring the quadrant where the teeth are located. ~\cite{ma2020srf} use the teeth feature vectors with neighborhood relations for better labeling accuracy, but it relies on a regular teeth distribution.

To emphasize the classification, the method proposes a new network and a postprocessing stage to fix potential errors. When classifying a single tooth in one patch, the network combines the shape features extracted from the patch with a tooth mask, and the position features indicating the patch's location on the model. Using the shape features or position features alone is not enough for accurate teeth classification. These two features are concatenated and fed into a fully-connected layer to obtain a 33-D output (32 teeth and one background). This is a more human-like way to address classification problems. The output is supervised by a cross-entropy loss. No prior information about whether the model belongs to the upper or lower jaw is needed in our method, because the incisors of the upper and lower jaw have the most significant difference in shapes and sizes, and the global feature extracted in this step will tell.

\noindent\textbf{Post-processing.}
Deep learning methods would not give the right label predictions every time, so a postprocessing stage using the dental arch curve to correct potential classification errors is proposed. First, predicted centroids are predicted onto $xOy$ plane and fit a para-curve as the dental arch curve. Then, based on the curve, the relative positions for each tooth are calculated, and some typical errors can be fixed. For example, if two teeth have the same label, then their labels can be corrected through the sorted label sequence; or if the labels are disordered, they can be reordered based on the order of teeth.
%
%
\subsection{TeethSeg team - Tudor Dascalu \etal}
\begin{figure}[!t]
\centering
\includegraphics[width=\textwidth]{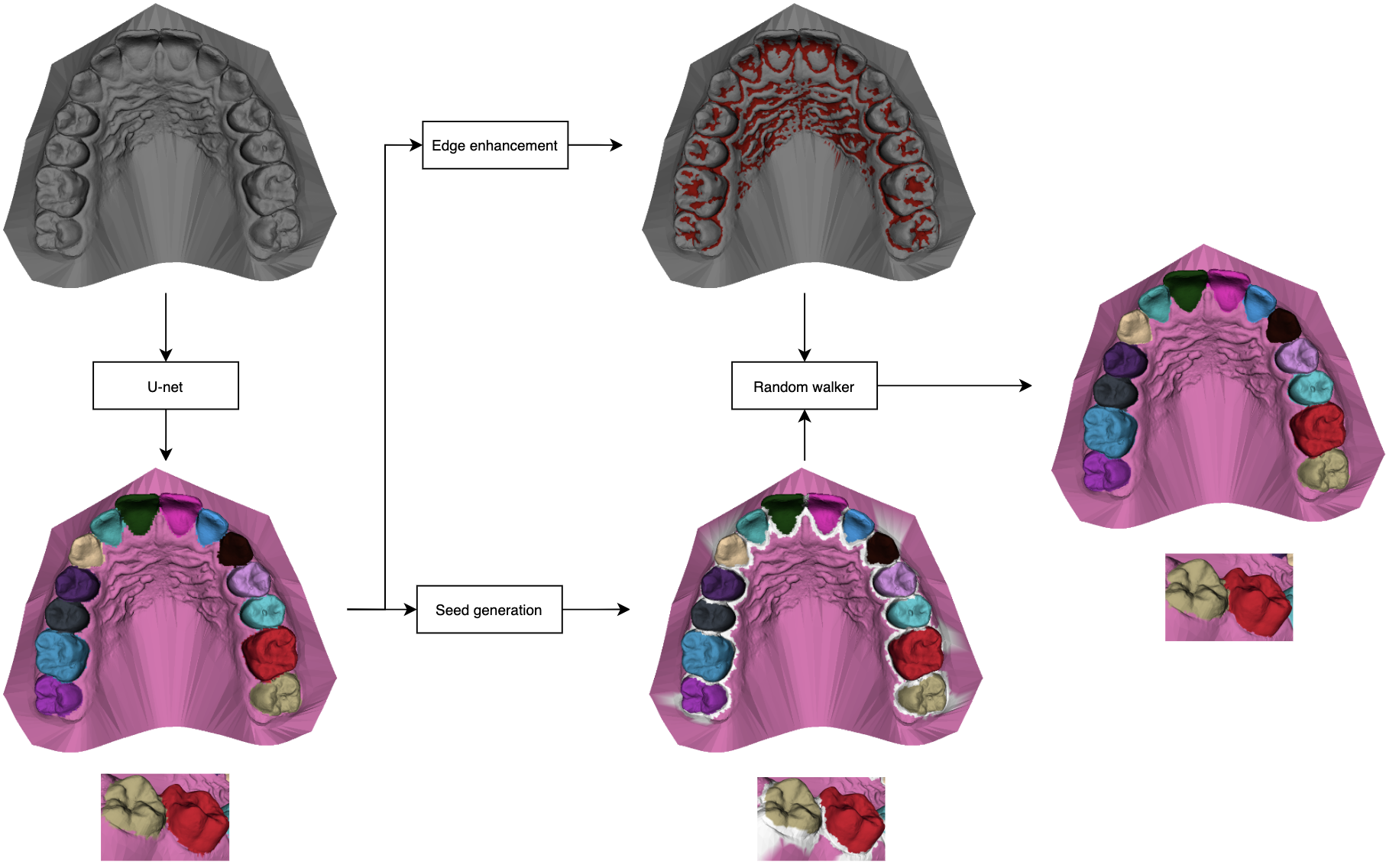}
\caption{Proposed approach by TeethSeg team. The mesh is segmented using the U-net model. The inaccuracies around tooth-tooth and tooth-gums edges are corrected using local geometric features.}
\label{fig:pipeline_teethseg}
\end{figure}

The team developed a multi-stage pipeline that aimed to label and segment teeth from 3D dental cast meshes (\figureabvr  \ref{fig:pipeline_teethseg}). The first step consisted of coarse segmentation of the dental structures. Next, the resulting outputs were refined using analysis of local geometric features. 
\subsubsection{Coarse segmentation}
The coarse segmentation phase consisted of indirect segmentation of the dental structures. The meshes were converted to binary volumetric images. Then, the resulting binary images featuring dental casts were segmented using the 3D U-net architecture. Separate U-net models were trained for the lower and upper jaws. Each model produced an N-dimensional 3D array of the same size as the input binary image of the dental cast. Individual teeth masks account for $N - 1$ of the channels in the array, while 1 mask corresponds to non-dental structures.
\subsubsection{Fine segmentation}
The fine segmentation stage was introduced in order to tackle the loss in accuracy around tooth-tooth and
tooth-gum edges, caused by the fact that the resolution of the binary images was much lower than the
resolution of the dental casts. To generate the fine segmentation,  the input dental cast was labeled
using the binary U-net result, by finding the closest grid point to a given mesh vertex. This labeling was coarse because of the aforementioned resolution discrepancy between the binary volumetric images and
the meshes. In the refinement stage, the participants defined a function $f$ that enhanced vertices close to edge regions. The output of the function $f$ for a given vertex $v$ was based on the convexity of the neighborhood surrounding $v$. Then, the vertices of the dental cast were labeled using the Random Walker algorithm. The seed points were set to vertices belonging to the dental crown and gum regions, extracted from the coarse segmentation output. For each non-seed vertex, random walkers started navigating the mesh space until they reached a seed point.  The movement of the walkers was steered using the function $f$, such that they were less likely to step over edge regions. The results of the Random Walker algorithm were used as the final outputs of the framework.
%
%
\subsection{OS team - Tae-Hoon Yong \etal}
\begin{figure}
\includegraphics[width=1\textwidth]{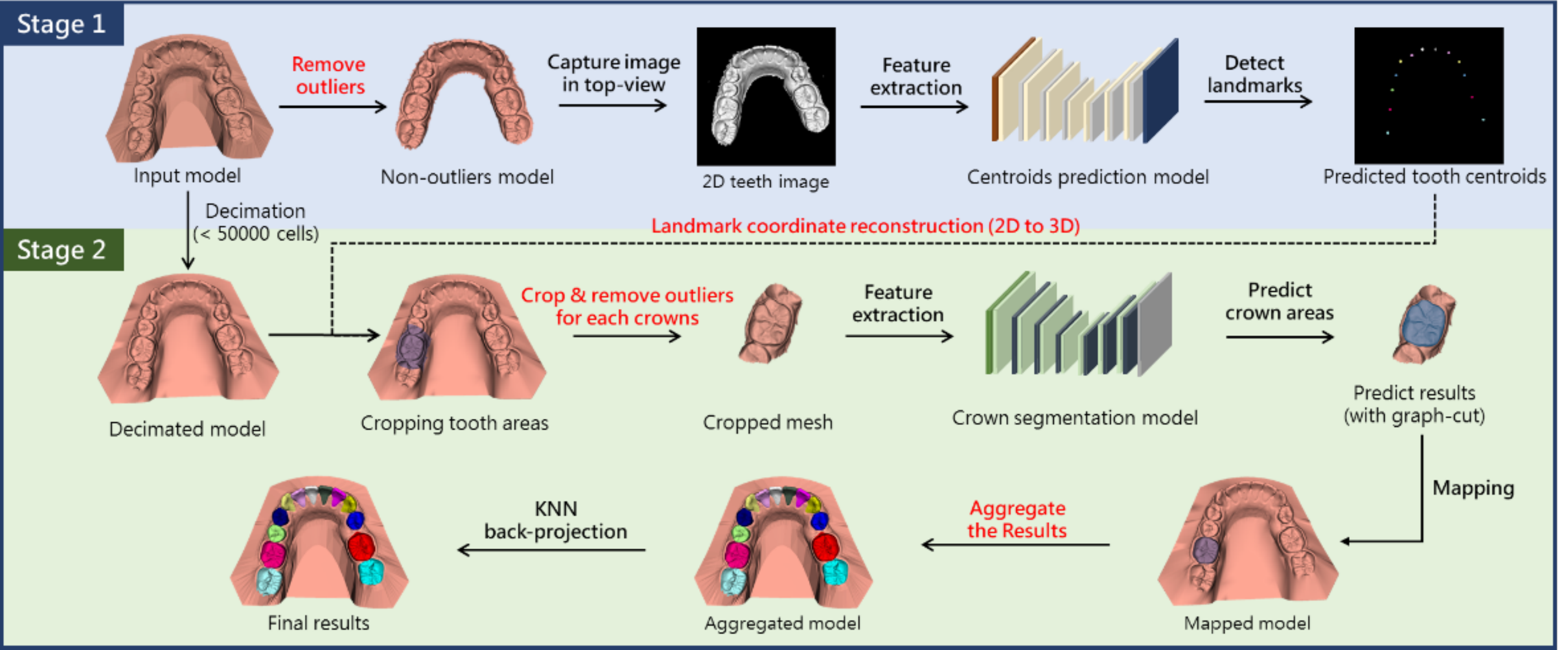}
\caption{Proposed approach by OS team. In stage 1, we obtained a 2D captured image from the input mesh  pre-processed to remove outliers. The deep learning model of the 2D-based encoder-decoder structure predicted the centroids and FDI numbers of each tooth from the 2D images. In stage 2, we acquired a cropped mesh for each tooth based on the results of stage 1 from a mesh decimated with less than 50,000 faces. After each crown region segmented from the cropped mesh through the crown segmentation deep learning model was restored and aggregated to the original mesh, aggregated model was back-projected to the original using the K-Nearest Neighbor (KNN) algorithm.}
\label{fig:osteam_pipline}
\end{figure}

As shown in Fig \ref{fig:osteam_pipline}, the proposed approach consists of two stages which combined teeth centroids prediction and classification based on the two dimensional image with Federation Dentaire Internationale (FDI) dental numbering system and individual teeth segmentation based on mesh surfaces. 

\subsubsection{Teeth centroids prediction and classification}
Given an input mesh model, we first extracted the outliers to obtain normalized model with some noise removed. To identify the centroids of each tooth, we obtained two dimensional teeth image captured from top view rendering which was perpendicular to the occlusal plane. Through the captured image with 512×512, we predicted the centroids of each tooth using a heatmap prediction method based on the high-resolution networks \citep{wang2020deep}, which were consisted of subnetworks in parallel with fusing low-to-high representations. In addition to detecting the centroids, each landmark was also classified with 16 channels. As the predicted centroids showed the clustering tendency, we applied DBSCAN algorithm \citep{schubert2017dbscan}  to all the predicted centroids for removing redundant tooth centroids. To restore the three dimensional coordinates closest to the two dimensional coordinates among the points located in the crown, the three dimensional coordinates were calculated by projecting the 3D coordinates into the two dimensional space. 
\subsubsection{Individual teeth segmentation}
Before cropping each tooth based on the stage 1 results, we decimated the number of faces to 50,000 or less from the input mesh model to compute efficiently. Based on the nearest adjacent centroid landmarks, we calculated the radius of individual teeth for cropping the mesh into an ellipse or circle shape. The circle shape was applied to the molars, and the ellipse shape was applied to the other teeth. To segment the tooth region in the cropped individual teeth faces, we used binary tooth segmentation methods improving the implementation of the graph-constrained learning module \citep{lian2020deep,wu2022two}. Based on the results of segmented individual teeth, we refined the results to obtain accurate crown areas using graph-cut algorithm \citep{lian2020deep}. After performing the refinement method on individual teeth area, we mapped each result into the decimated model and aggregated it. Finally, the decimated models were up-sampled to the original scan model using the K-Nearest Neighbor (KNN) algorithm  \citep{johnson2019billion}, which preserved the results of the decimated model.
%
%
\subsection{Champers team - Niels van Nistelrooij \etal}
\subsubsection{3D Scan pre-processing}
An intra-oral scan was represented by its vertices with the coordinates and normals as vertex features. During pre-processing, the vertices were subsampled by overlaying a regular grid and sampling one point from each occupied grid cell. This resulted in a uniform-density point cloud with a variable number of points that remains to scale. Random data augmentation scaled the coordinates, flipped the scan left to right, and rotated it around the longitudinal axis.
The annotations on the source scan were used to determine the centroid and FDI label of each tooth instance. The 32 possible labels were subsequently translated to 7 classes by removing the distinction of upper/lower and left/right. Furthermore, the few third molars were translated to second molars.

\subsubsection{Proposed model architecture}
The proposed model is largely inspired by TSegNet \citep{cui2020tsegnet}, which split the problem into two stages for centroid prediction and tooth segmentation, respectively.
\paragraph{Encoder-Decoder Network}

An encoder-decoder architecture that was repeatedly used through-
out the model was the Stratified Transformer \citep{stratified_transformer}. This Vision Transformer for point clouds uses window-based multi-head self-attention to learn contextual features. Its encoder path uses shifted windows in subsequent Transformer blocks to efficiently learn long-range dependencies \citep{swin_transformer}. After each set of blocks, the current point cloud was subsampled with Farthest Point Sampling. The initial point to seed Farthest Point Sampling was sampled randomly, such that the subsampled point cloud at the end of the encoder path was different for each forward pass.

In the decoder path, a smaller point cloud was upsampled using weighted 3-nearest neighbour interpolation. Its features were then summed to the features of a point cloud following a skip connection from the encoder path. The decoder path resulted in the same point cloud as the input, but with contextual features.

\paragraph{Prediction of tooth centroids}

First, the input point cloud was processed by the encoder path of a Stratified Transformer. Each point of the resulting subsampled point cloud had 256 features. These features were further processed by two fully-connected heads.

The first head predicted the Euclidean distance from each point to its closest tooth centroid and was supervised with a smooth L1 loss. These distances were subsequently used to filter points in the periphery, such that only points on the dental arch were retained.

The second head predicted the x-, y-, and z-offsets from each point to its closest tooth centroid. These predictions were compared to the ground truth with the following loss function:

\begin{equation*}
  \label{eq:loss_final}
  \mathcal{L}_{CP} = \overbrace{\frac{1}{K} \sum_{i=1}^N \frac{1}{r_i^{\scaleto{(1)}{5pt}}} ||\hat{p}_i + \hat{o}_i - c_i^{\scaleto{(1)}{5pt}}||^2_2}^{\text{Normalized Euclidean}}
  \quad + \quad
  \overbrace{\frac{1}{K} \sum_{i=1}^N \frac{r_i^{\scaleto{(2)}{5pt}}}{r_i^{\scaleto{(1)}{5pt}}} \frac{||\hat{p}_i + \hat{o}_i - c_i^{\scaleto{(1)}{5pt}}||_2}{||\hat{p}_i + \hat{o}_i - c_i^{\scaleto{(2)}{5pt}}||_2}}^{\text{Separation}}
\vspace{-0.3cm}
\end{equation*}
where
\begin{itemize}
  \item $K$ is the number of ground-truth tooth centroids;
  \item $\hat{p}_i \in \mathbb{R}^3$ are the coordinates of a subsampled point after filtering;
  \item $\hat{o}_i \in \mathbb{R}^3$ are the corresponding predicted offsets;
  \item $c_i^{\scaleto{(k)}{5pt}} \in \mathbb{R}^3$ is the $k^{\text{th}}$ closest ground-truth tooth centroid from $\hat{p_i}$; and
  \item $r_i^{\scaleto{(k)}{5pt}} \in \mathbb{R}$ is the radius of the tooth instance of which $c_i^{\scaleto{(k)}{5pt}}$ is the centroid.
\end{itemize}

The Normalized Euclidean function computes the squared Euclidean distance between a point and its closest ground-truth tooth centroid. A bias in favor of larger teeth was remedied by normalizing by the radius of the tooth instance which has the closest centroid. Furthermore, the ground-truth centroid was chosen from the point's position, not from the point's predicted closest centroid.

The Separation function computes the difference in distances from a point to its closest and second-closest ground-truth centroids. This punishes the model whenever it predicts a further centroid. The normalization and separation resulted in a much better stratification of the predicted centroids among the tooth instances.

The Centroid Prediction stage was run six times to collect many centroid predictions. Finally, the DBSCAN algorithm was applied to cluster the centroids to one point per tooth instance \citep{dbscan}.

\paragraph{Tooth Segmentation}

The first step processed the input point cloud by a Stratified Transformer, which resulted in the same point cloud with 48 features for each point. For each predicted centroid, 3,528 nearest neighbours were sampled to form a proposal. Each point of such a proposal had 52 features; 48 contextual features, 3 global coordinates, and the local distance from the predicted centroid to the point.

Each proposal was then independently processed by a cascade of two Stratified Transformers. The first network predicted the binary segmentation of the foreground tooth in the proposal. This 1-channel prediction was channel-wise concatenated to form proposals with points that have 53 features.

These proposals were subsequently processed by the second network, which also predicted the foreground segmentation. The predictions of both networks were supervised by the binary cross-entropy loss.

The second network additionally predicted the position of the foreground tooth, provided the segmentation from the first network. In order to do that, it first applied global average pooling to the features in the most latent dimension, followed by fully-connected layers to return seven logits per proposal. This classification was supervised with the categorical cross-entropy loss. Because the classification head converged much faster compared to the segmentation heads, its learning rate was reduced by a factor of ten.

The Tooth Segmentation stage was run twice to collect multiple proposal predictions per tooth instance. For each run of the Tooth Segmentation stage, the same predicted centroids were used.
\subsubsection{Post-processing}
First, pairs of proposals were merged whenever their foreground points had an intersection-over-union of at least 0.35. Segmentation and class logits from points that occurred in both proposals were summed. This naturally merged proposals from multiple runs and proposals with the same foreground tooth, thereby increasing effectiveness.

Then, the segmentations were interpolated to the points of the source point cloud. To this effect, 20\% nearest neighbours were sampled around each predicted centroid. The segmentation logits were then piece-wise linearly interpolated to the source points, falling back to 3-nearest neighbors for extrapolation. The classification logits were thus not interpolated.

The next step translated the predicted positions back to FDI labels. Whether the jaw was an upper or lower jaw was prior information and the left/right distinction was made by comparing a predicted centroid to the centroids of central incisor proposals. The third molars were retrieved by incrementing the label of a second molar if there was an additional anterior second molar next to it.

Finally, the interpolated tooth proposals were projected back to the source point cloud. Source points not present in any proposal or with only negative segmentation logits were attributed the gingiva label. Other source points were attributed the FDI label of the proposal that gave the point the highest segmentation logit. Furthermore, to allow for multiple tooth instances with the same FDI label within a scan, the index of the attributing proposal was also returned.

\subsubsection{Implementation Details}

The Centroid Prediction stage was trained for 500 epochs on 1191 scans from the challenge; 7 scans with braces were removed, as well as 2 scans with severely misaligned teeth.

The Tooth Segmentation stage was trained for 100 epochs on 1072 scans, where 119 scans were left for model selection. This 90\%/10\% split was determined with a second-order stratification based on the unique FDI labels of each scan \citep{scikit_multilearn}.

The AdamW optimizer was used with weight decay, as well as a cosine annealing learning rate scheduler with a linear warmup period \citep{adamw}. Lastly, all models were implemented from the ground up using PyTorch Lightning \citep{pytorch}.
\section{Experimental results}
We present the results under quantitative and qualitative evaluation. We first present the final ranking of the participating methods as well as the obtained results for the different tasks of the challenge. Afterword we expose different cases to visually check the quality of the predictions obtained for the final phase of the challenge. 

\subsection{Quantitative evaluation}

The ranking shown in Table \ref{tab:quantitative_evaluation} is determined by the global score previously described in Section \ref{sec:materials}. It is worth noting that the ranking may differ depending on the specific task or metric being evaluated. In terms of overall performance, the method proposed by the CGIP team holds the top position. However, when focusing specifically on the teeth localization task, the FiboSeg team achieves the highest score with an Exp(-TLA) of 0.9924. On the other hand, the CGIP team demonstrates exceptional performance in teeth segmentation task, obtaining a TSA score of 0.9859. The IGIP team exhibits the best performance in the tooth labeling task, achieving a TIR score of 0.9289. These results emphasize the diversity and strengths of the different methods, showcasing their effectiveness in specific aspects of the challenge.

\begin{table}[h]
    \centering
    \begin{tabular}{lllll}
    \toprule
    \rowcolor{white!40}
    \textbf{Team} & \textbf{Exp(-TLA)} & \textbf{TSA} & \textbf{TIR} & \textbf{Score} \\
    \midrule
     CGIP & 0.9658 & \textbf{0.9859} & 0.9100 & \textbf{0.9539} \\
     FiboSeg & \textbf{0.9924} & 0.9293 & 0.9223 & 0.9480 \\
     IGIP & 0.9244 & 0.9750 & \textbf{0.9289} & 0.9427 \\
     TeethSeg & 0.9184 & 0.9678 & 0.8538 & 0.9133 \\
     OS & 0.7845 & 0.9693 & 0.8940 & 0.8826 \\
    Chompers & 0.6242 & 0.8886 & 0.8795 & 0.7974 \\
    \bottomrule
    \end{tabular}
    \caption{Obtained evaluation metrics for the participating teams. The given ranking is based on the final score (see last column). }
    \label{tab:quantitative_evaluation}
\end{table}
\subsection{Qualitative evaluation}
\begin{figure}[htbp]
\centering
\begin{tabular}{ccccc}
        & (a) & (b) & (c) & (d) \\

        \rotatebox{90}{Ground Truth} &
    \includegraphics[width=0.22\textwidth]{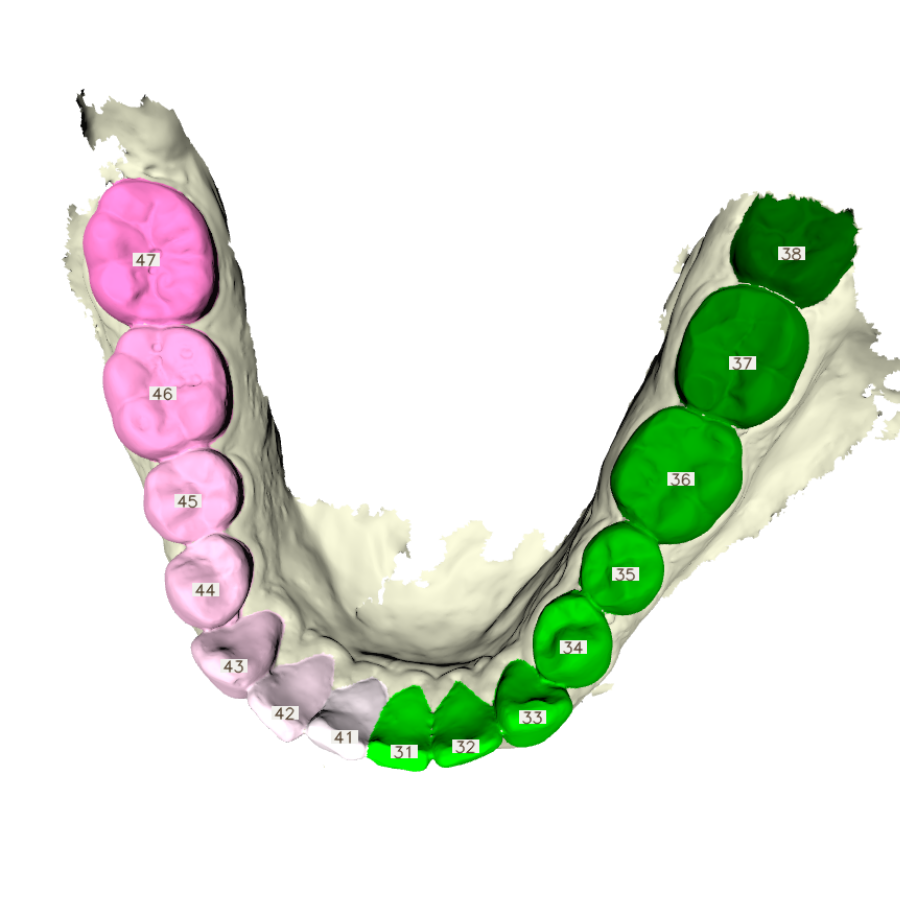} & 
    \includegraphics[width=0.22\textwidth]{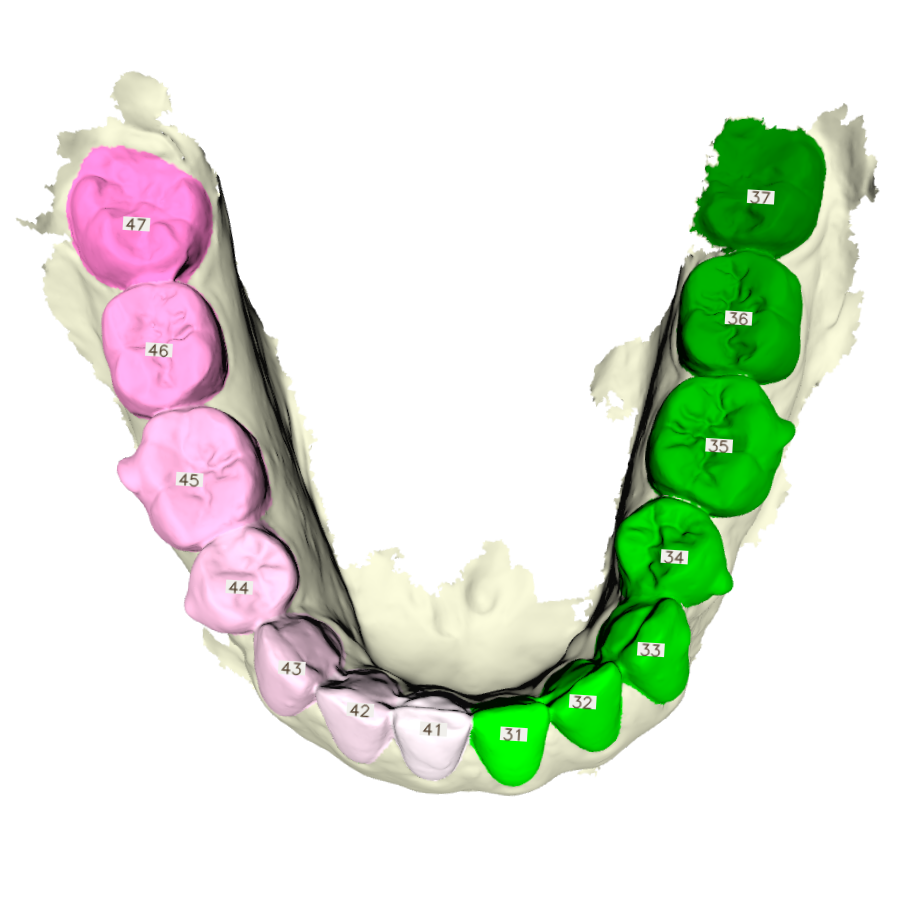} & 
    \includegraphics[width=0.22\textwidth]{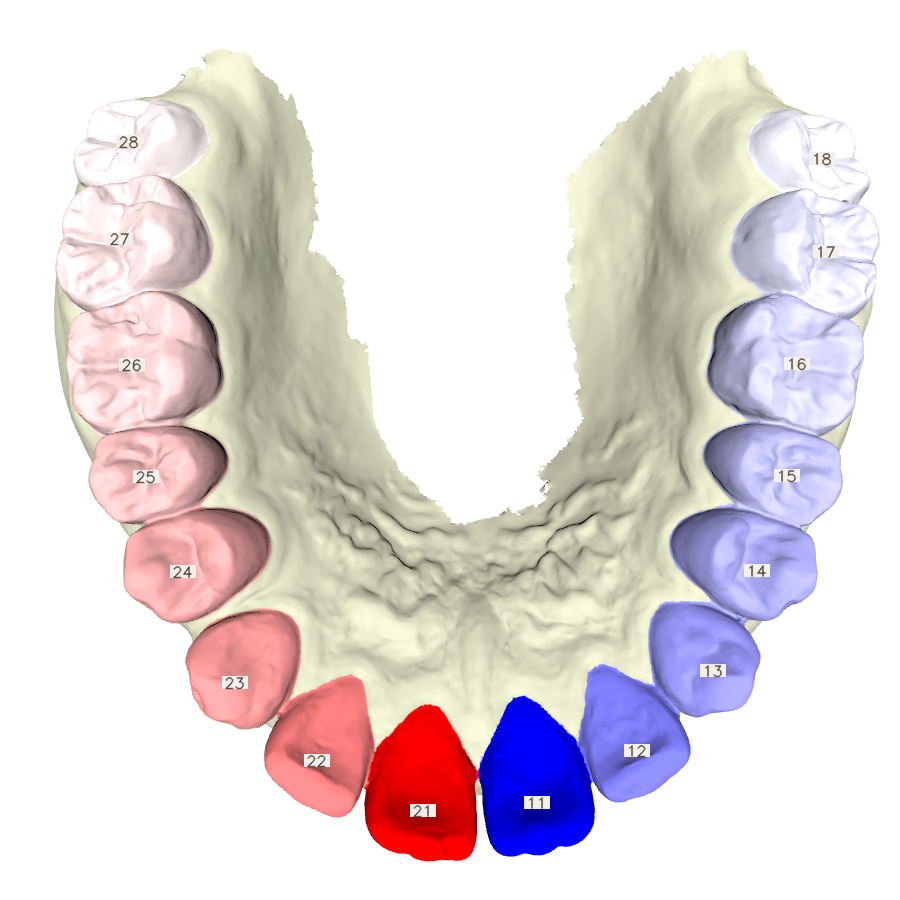} & 
    \includegraphics[width=0.22\textwidth]{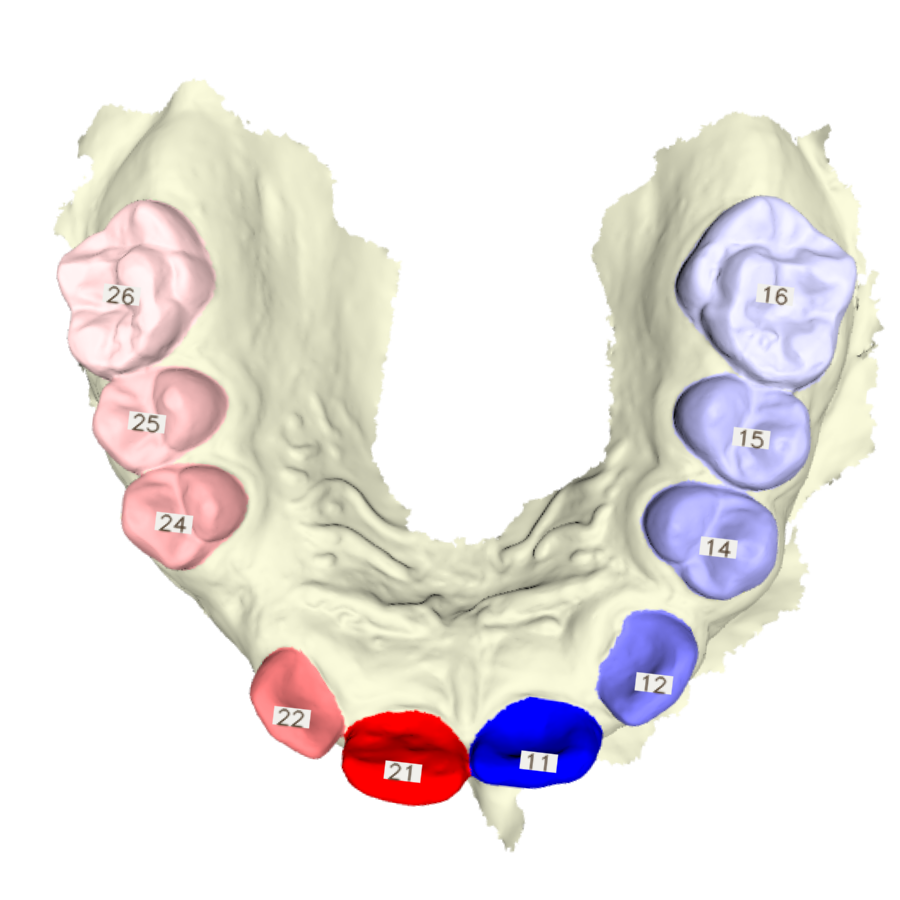} \\
        \rotatebox{90}{CGIP team} &
    \includegraphics[width=0.22\textwidth]{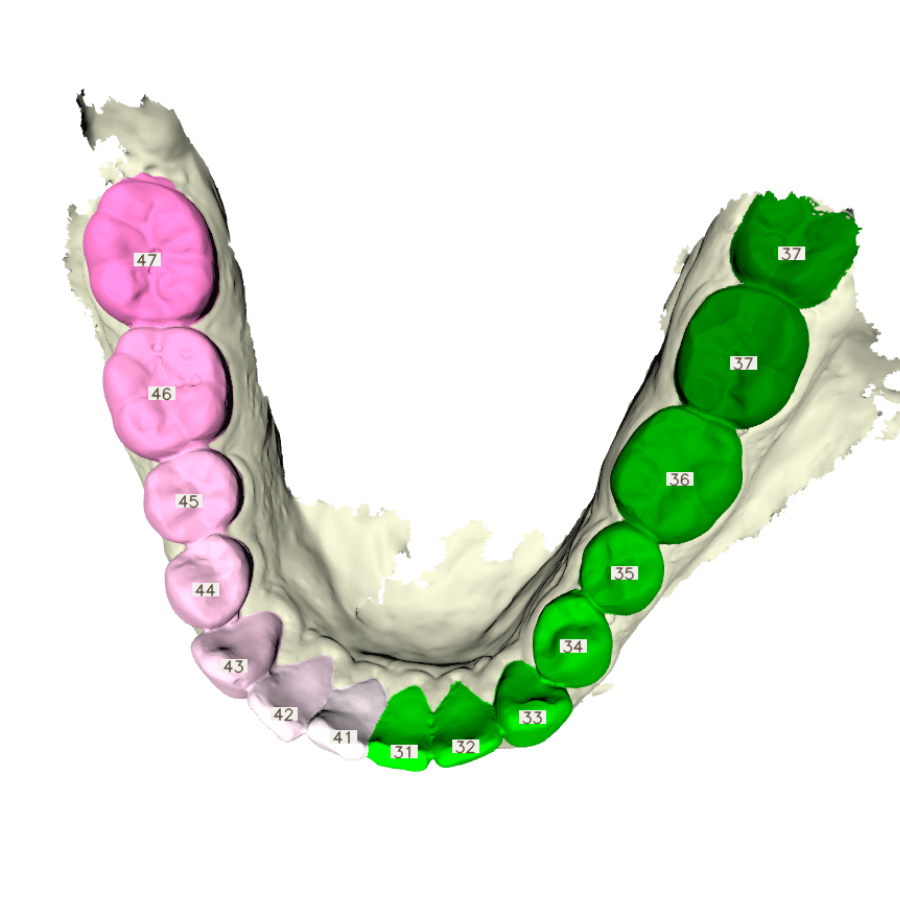} & 
    \includegraphics[width=0.22\textwidth]{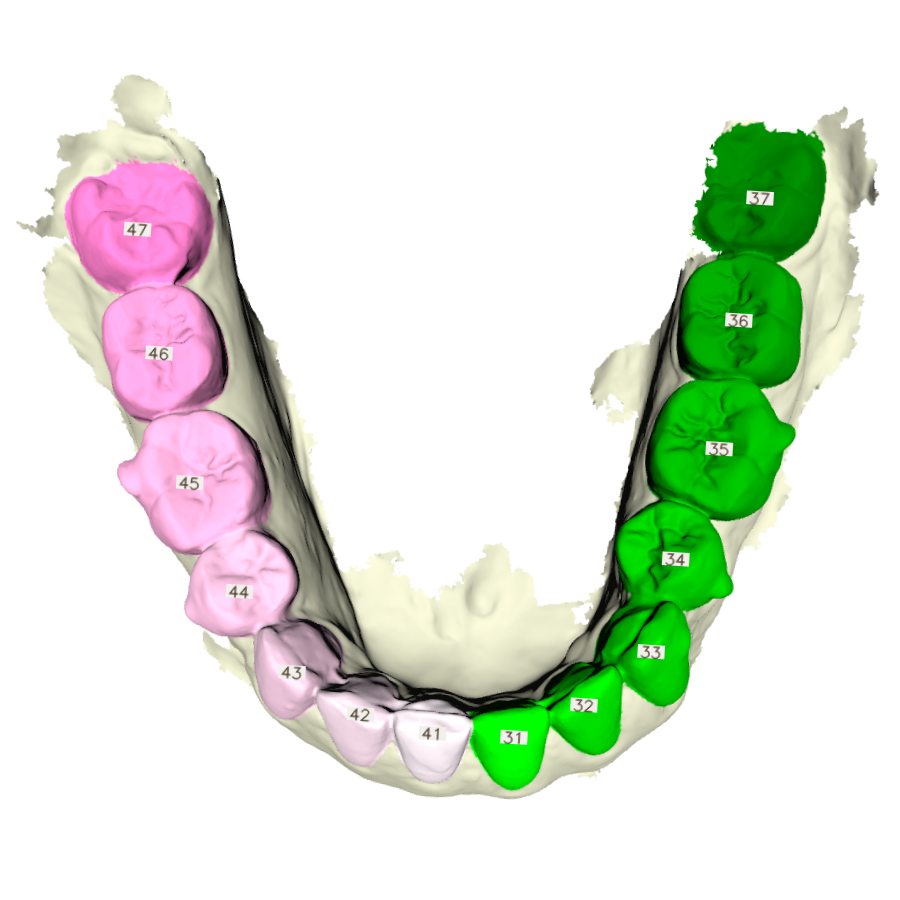} & 
    \includegraphics[width=0.22\textwidth]{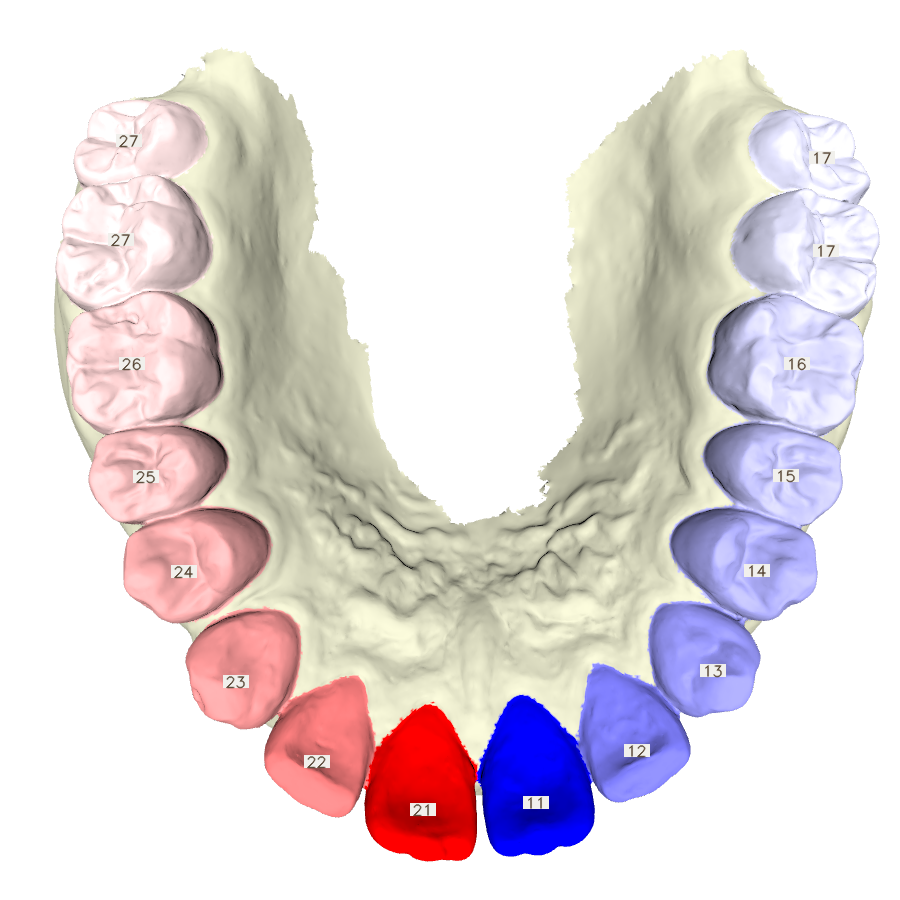} & 
    \includegraphics[width=0.22\textwidth]{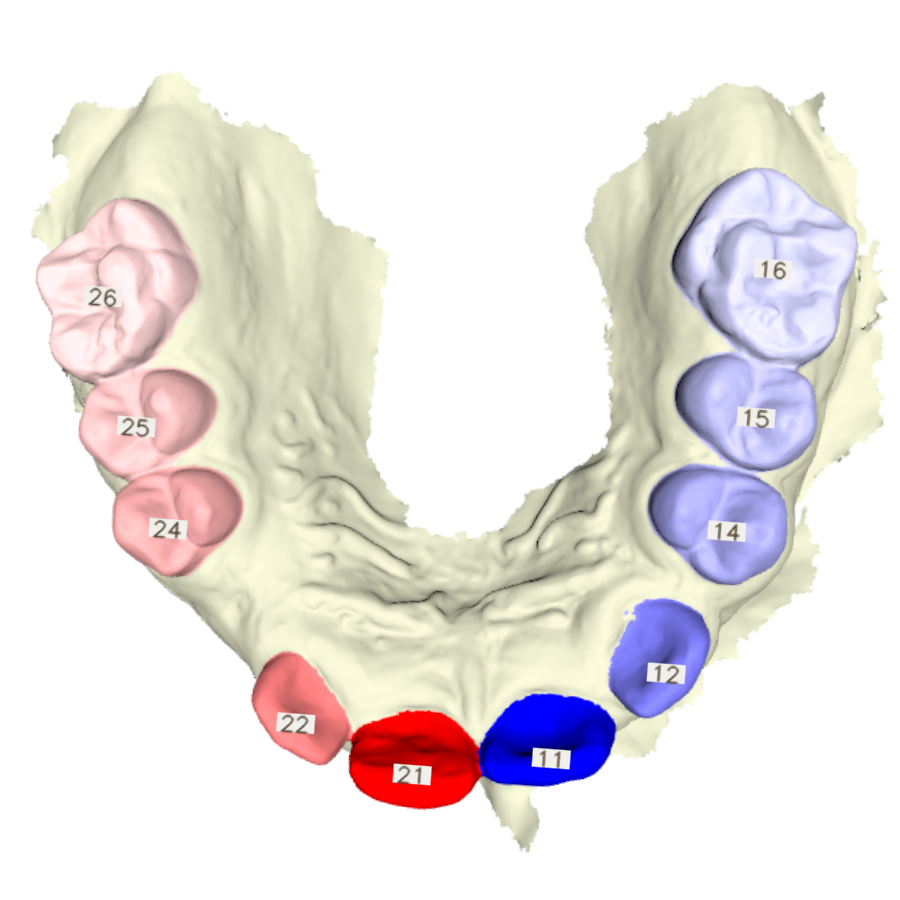} \\
        \rotatebox{90}{FiboSeg Team} &
    \includegraphics[width=0.22\textwidth]{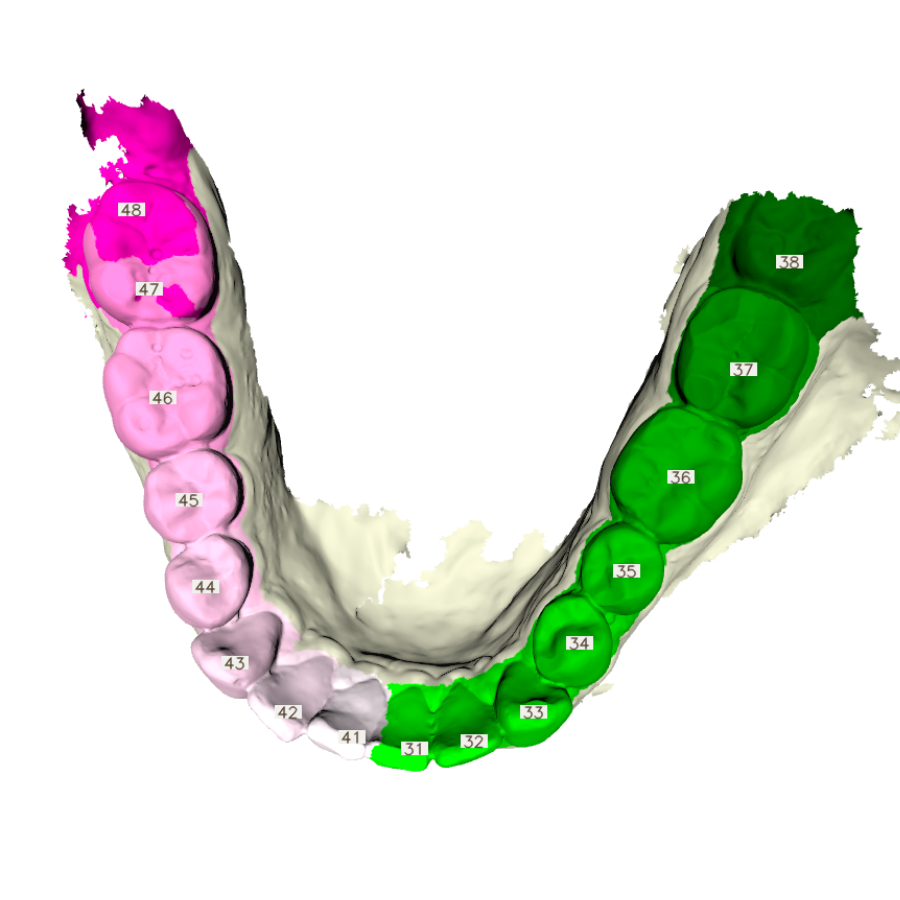} & 
    \includegraphics[width=0.22\textwidth]{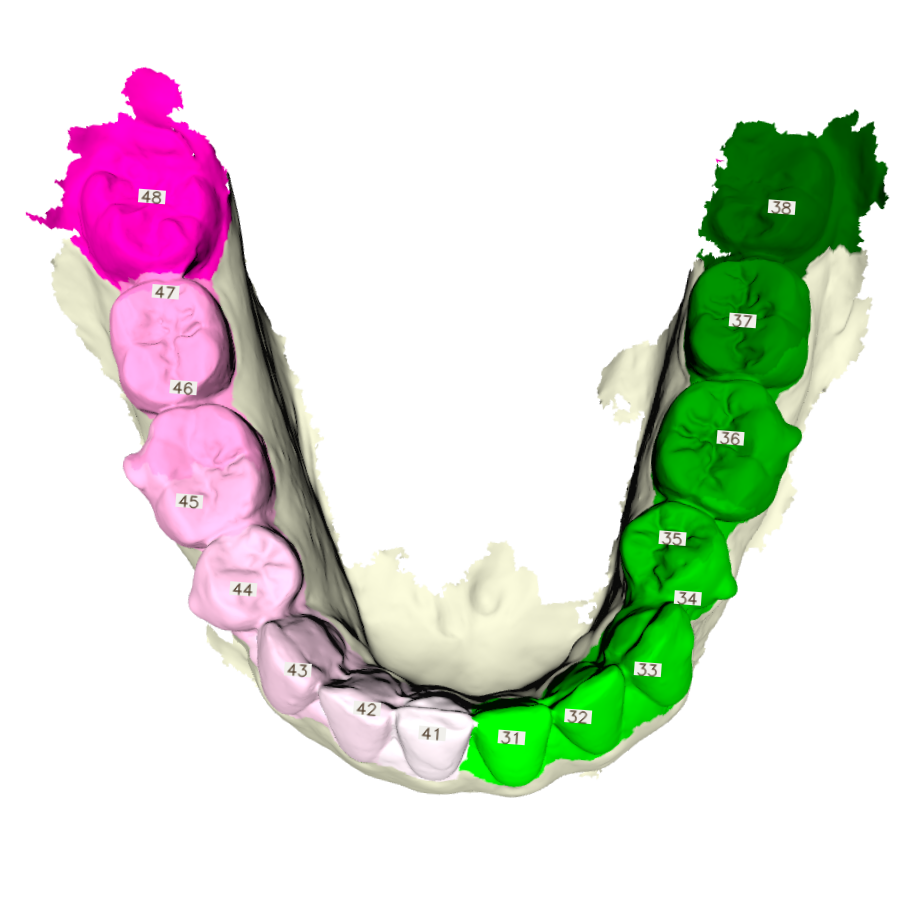} & 
    \includegraphics[width=0.22\textwidth]{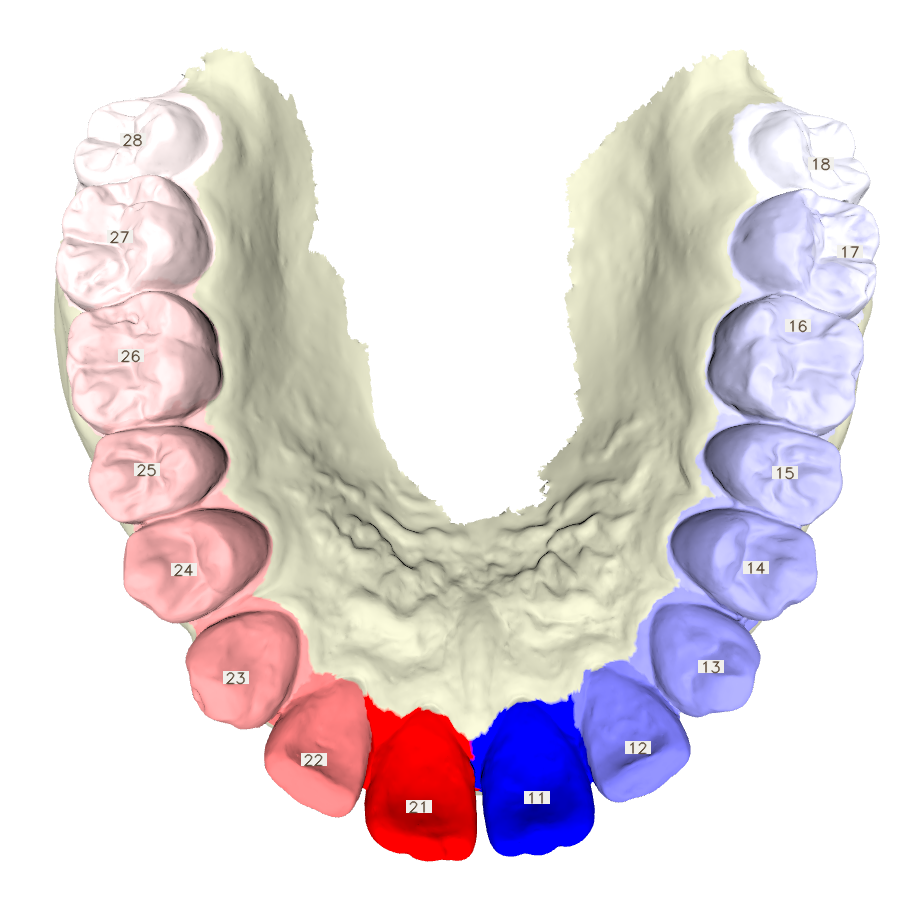} & 
    \includegraphics[width=0.22\textwidth]{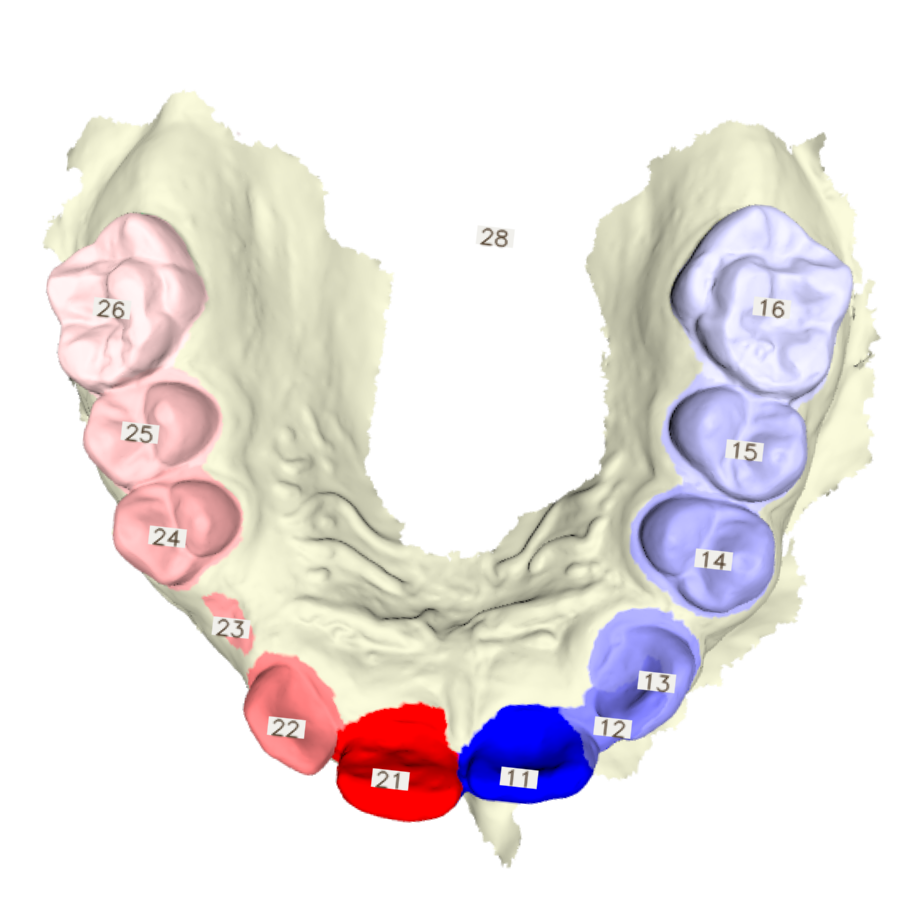} \\
        \rotatebox{90}{IGIP Team} &
    \includegraphics[width=0.22\textwidth]{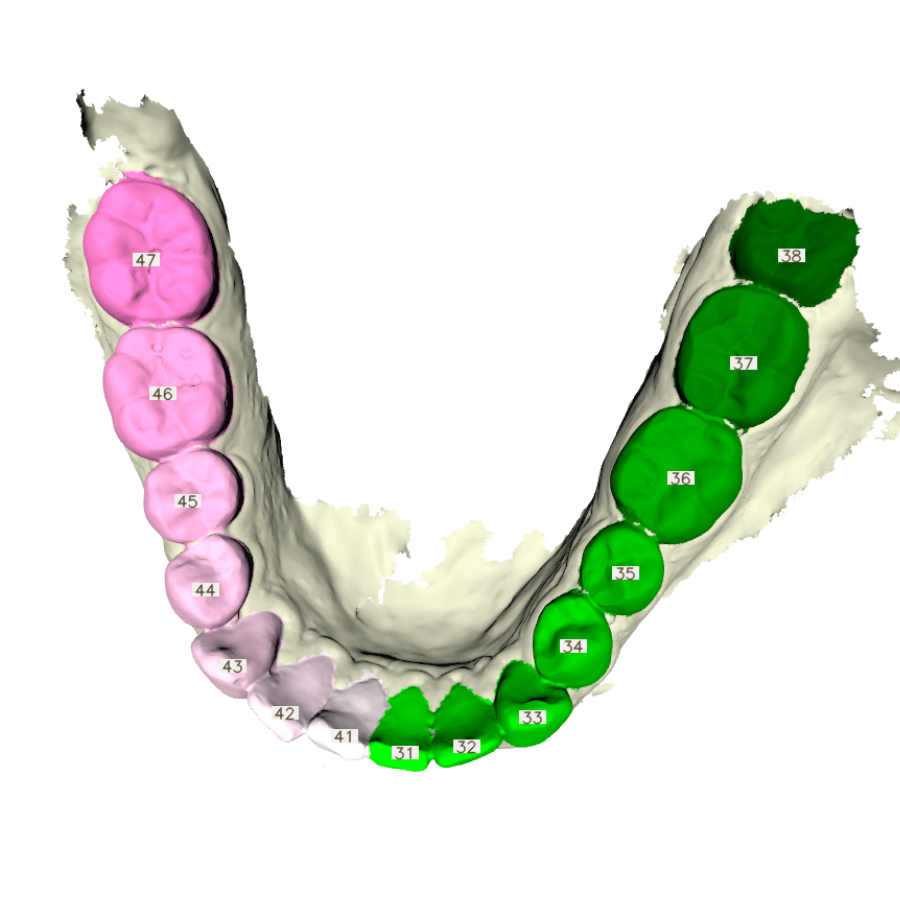} & 
    \includegraphics[width=0.22\textwidth]{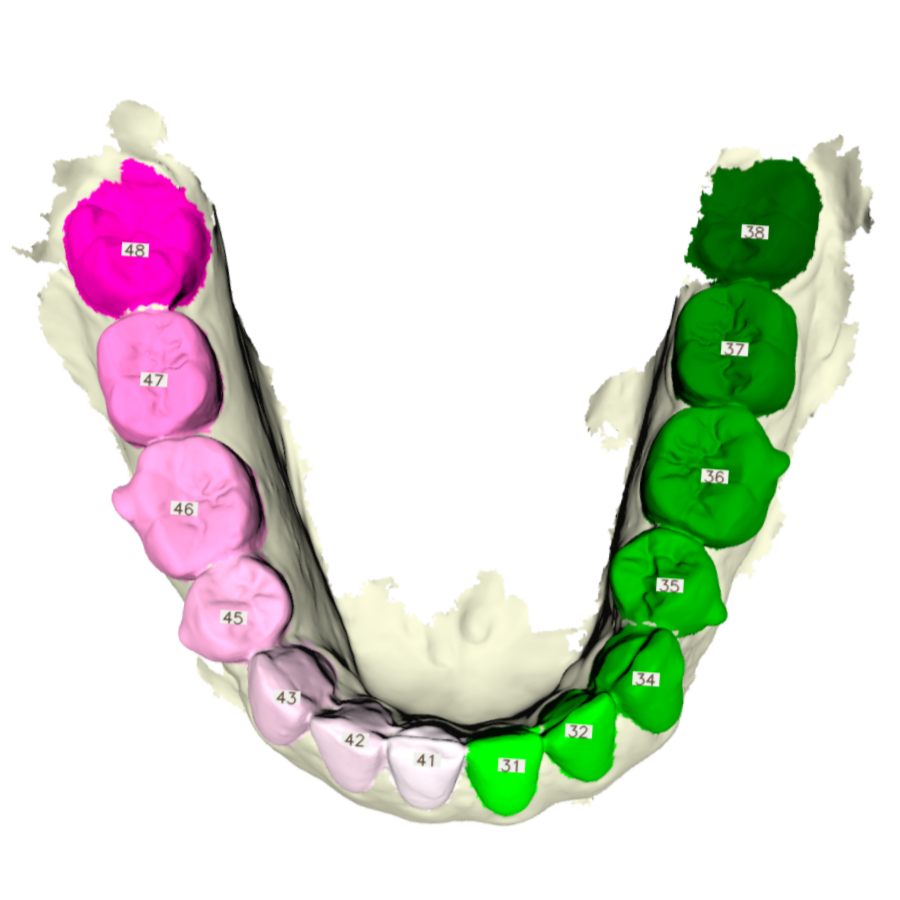} & 
    \includegraphics[width=0.22\textwidth]{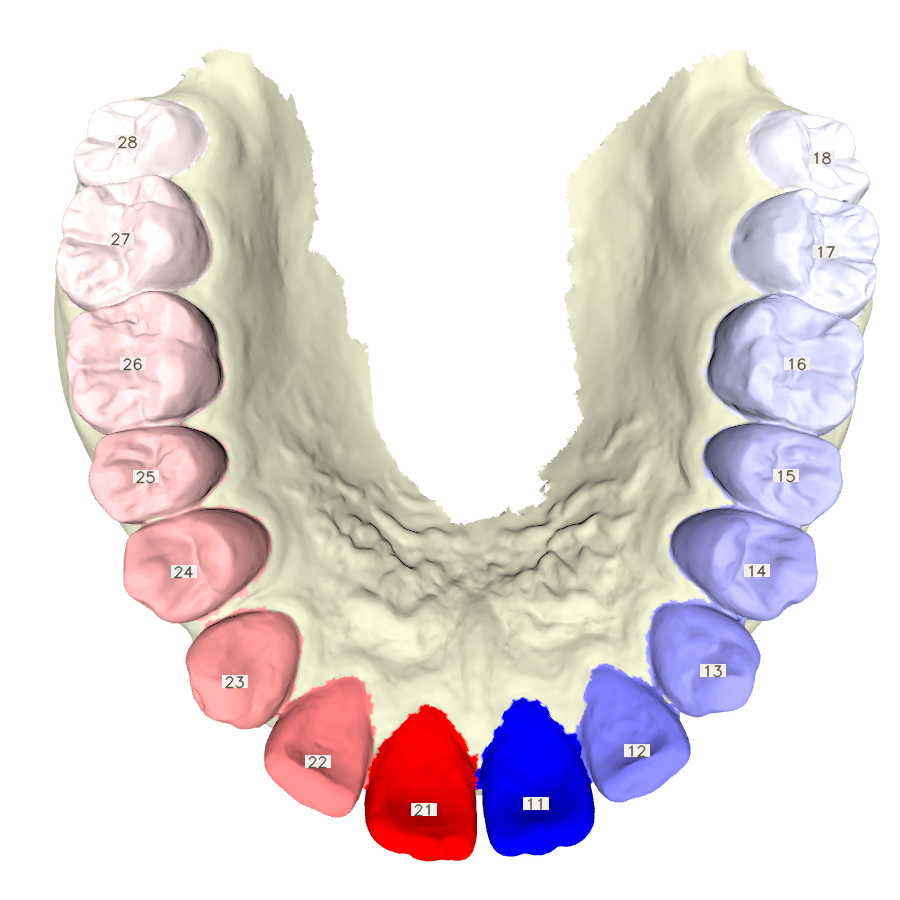} & 
    \includegraphics[width=0.22\textwidth]{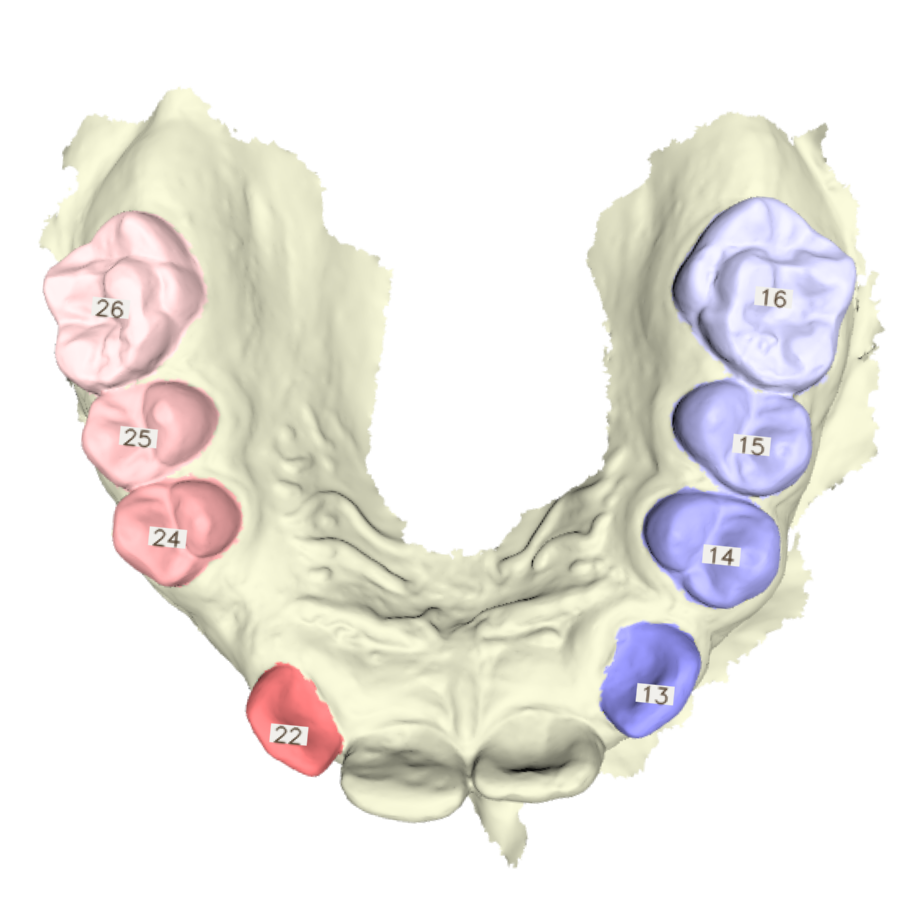} \\
        \rotatebox{90}{TeethSeg Team} &
    \includegraphics[width=0.22\textwidth]{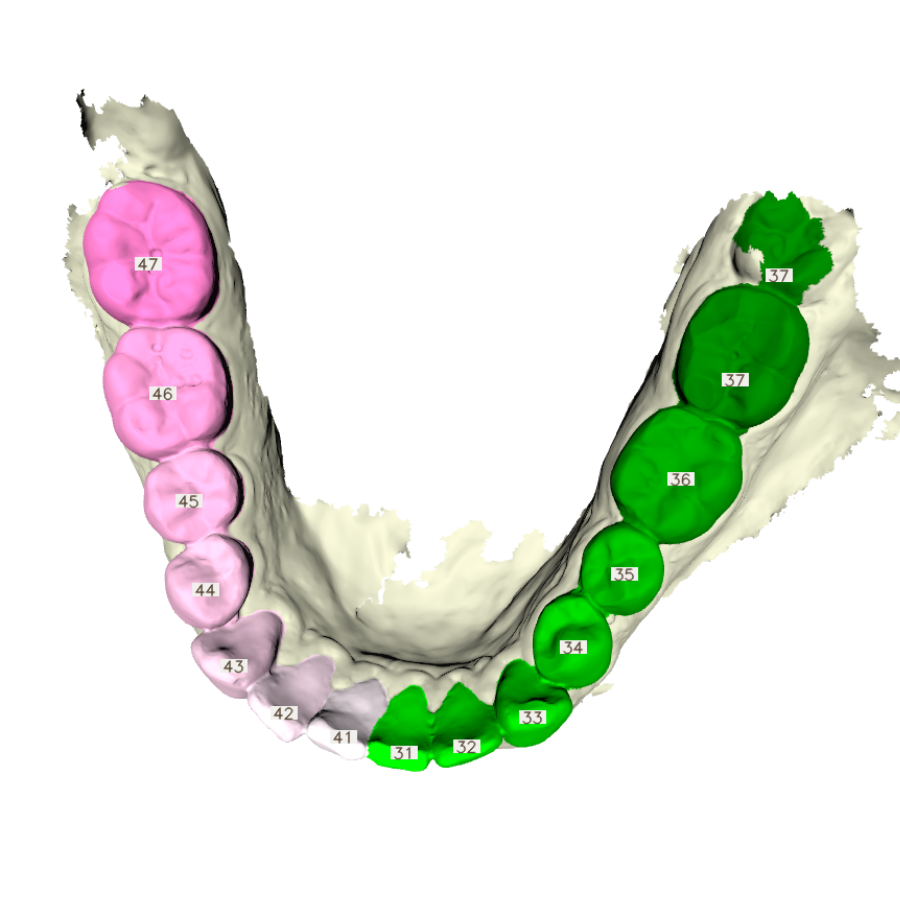} & 
    \includegraphics[width=0.22\textwidth]{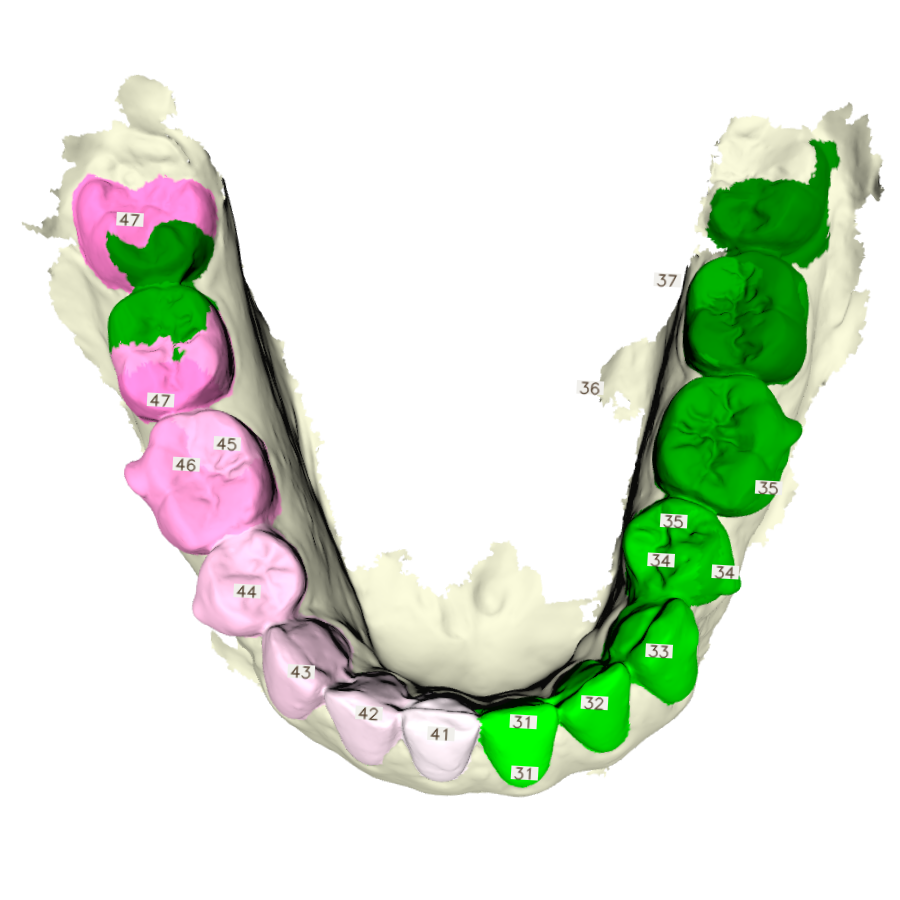} & 
    \includegraphics[width=0.22\textwidth]{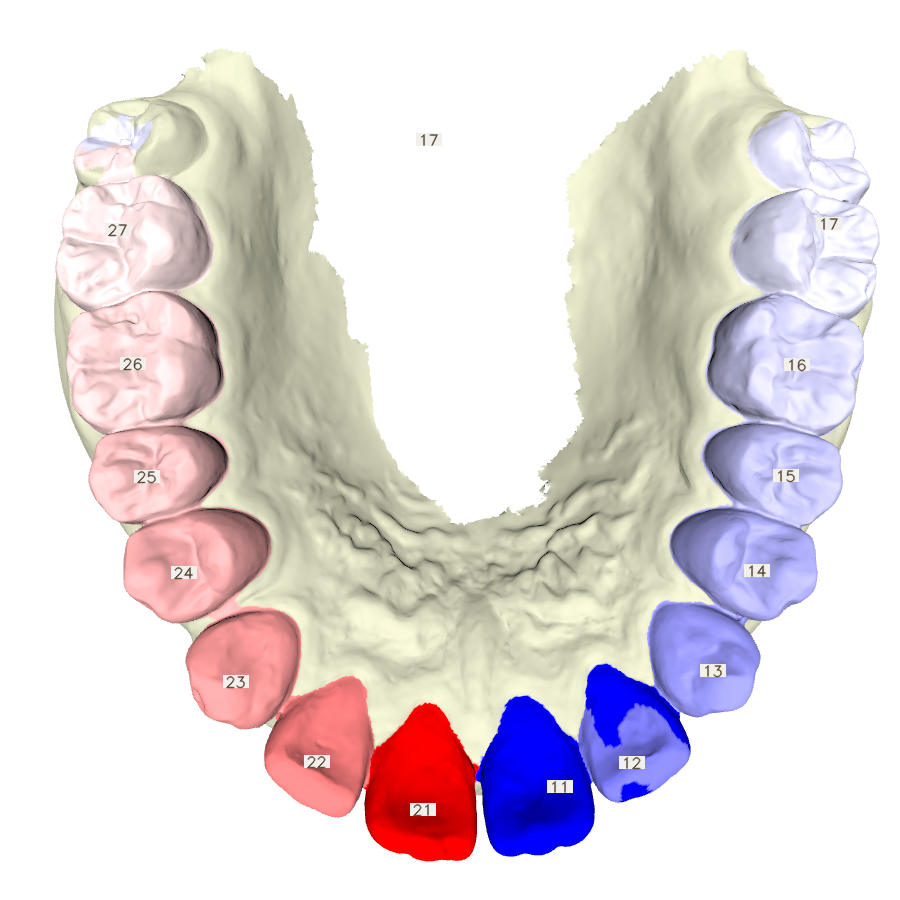} & 
    \includegraphics[width=0.22\textwidth]{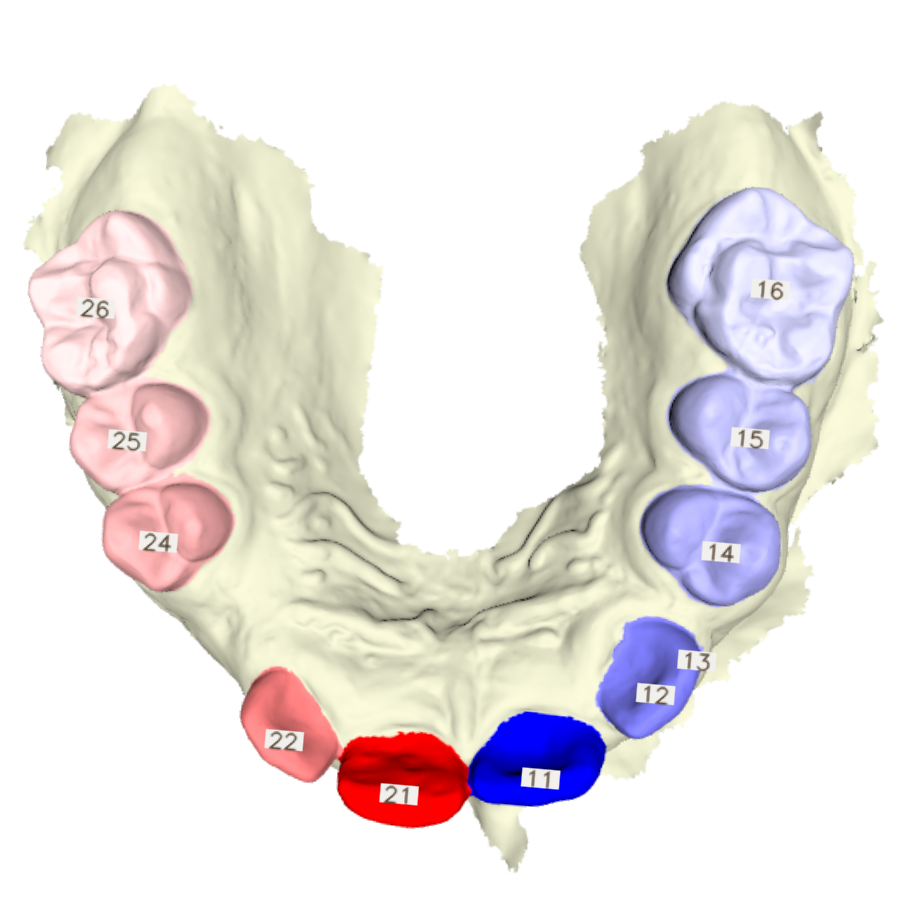} \\
        \rotatebox{90}{OS Team} &
    \includegraphics[width=0.22\textwidth]{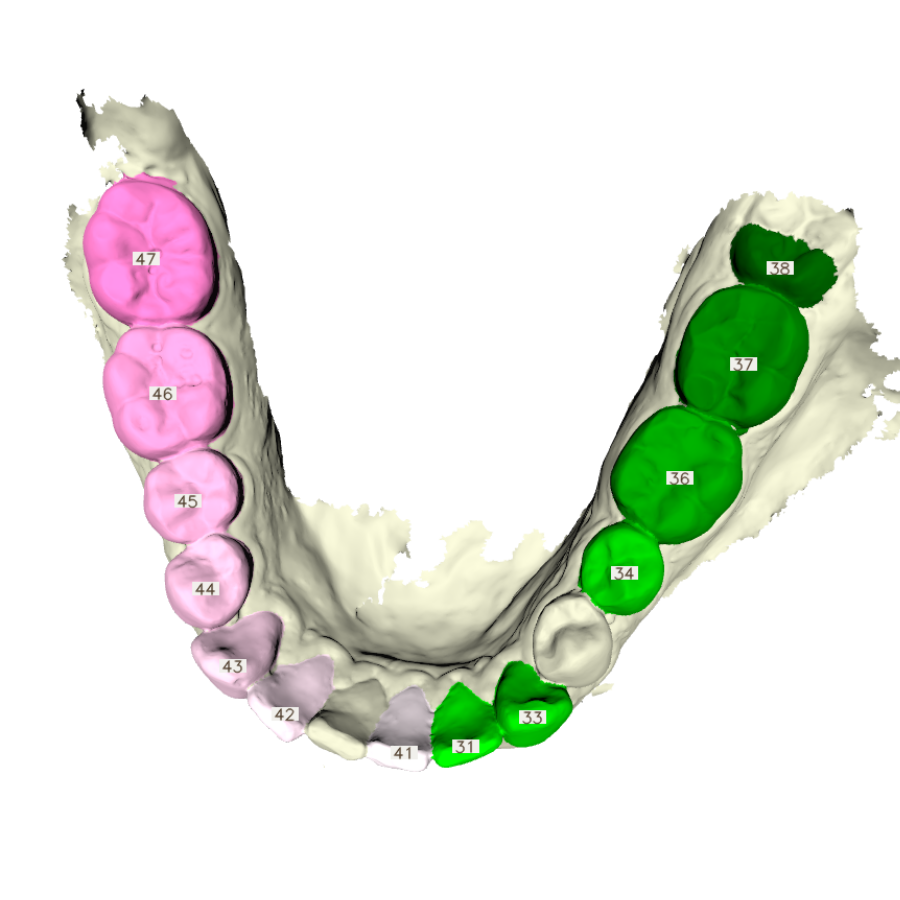} & 
    \includegraphics[width=0.22\textwidth]{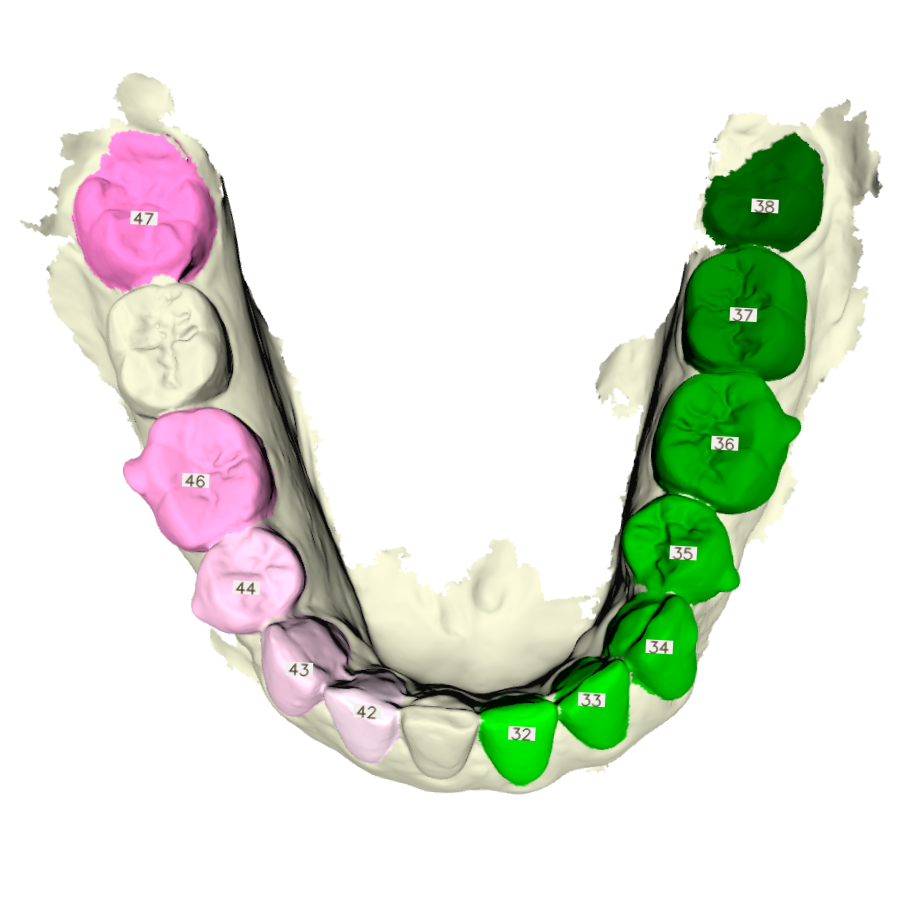} & 
    \includegraphics[width=0.22\textwidth]{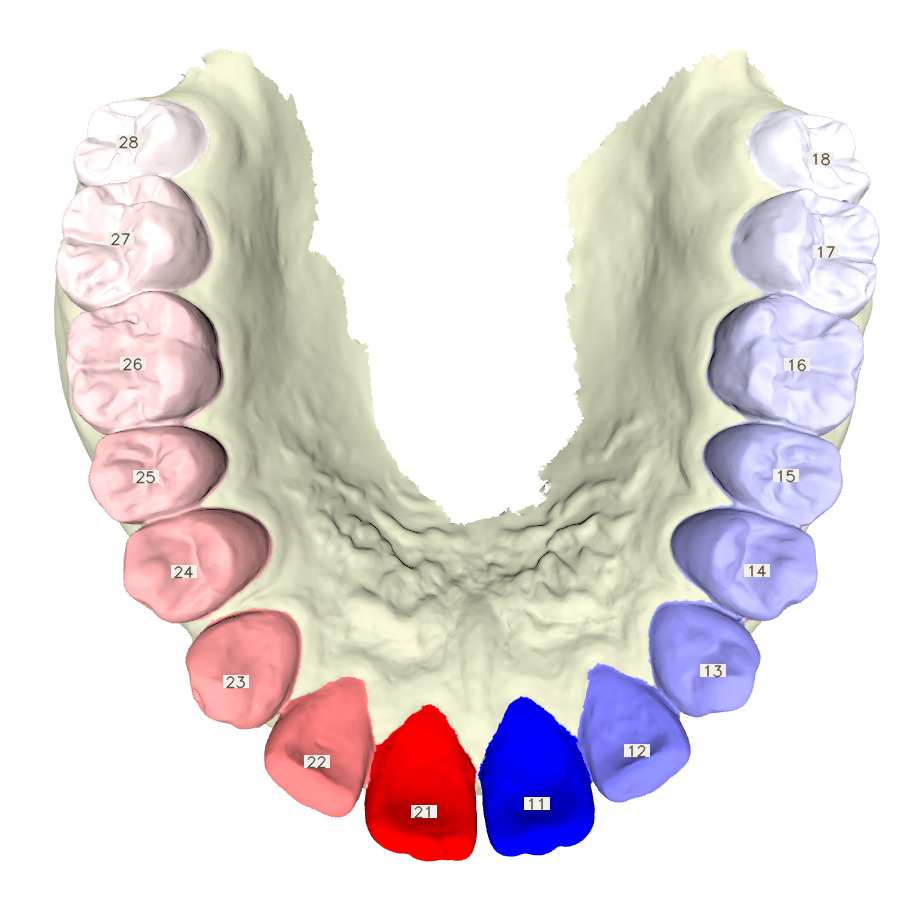} & 
    \includegraphics[width=0.22\textwidth]{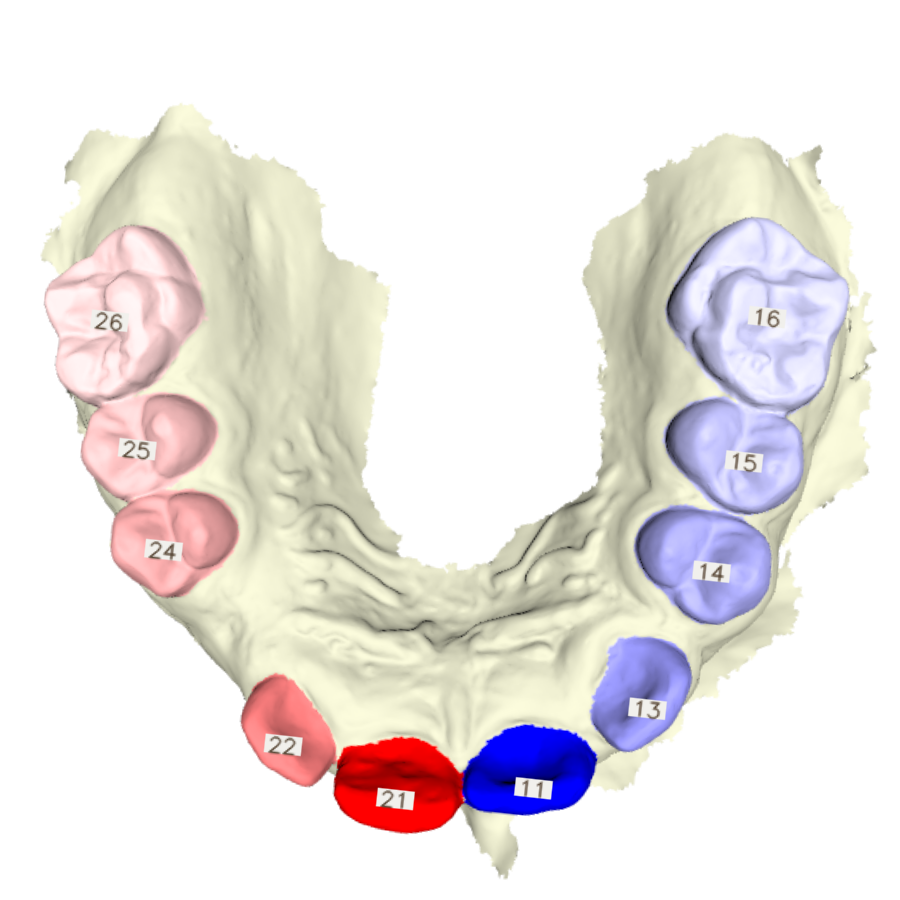} \\
\end{tabular}
\caption{Visual comparison of the obtained results by applying the competing methods on four samples, \ie two lower (a,b) and two upper (c,d) jaws. The first row shows the ground truth related to teeth segmentation and labeling. The remaining rows show the results following the team global ranking.}
\label{fig:qualitative_evaluation}
\end{figure}

Figure \ref{fig:qualitative_evaluation} provides a visual representation of the segmentation results obtained by the competing methods on four samples from the validation dataset. Overall, the visual evaluation of the obtained results aligns with the ranking provided in the quantitative evaluation. The CGIP team demonstrates superiority, particularly in the segmentation task, with consistently accurate segmentation results. However, it should be noted that the FiboSeg team exhibits lower segmentation accuracy, specifically in the gum-teeth border in most of the segmented teeth. Missing tooth detection is observed across multiple teams, but it is more pronounced in the results of the IGIP team (sample d) and the OS team (samples a and b, for instance). In the case of the TeethSeg team, there are instances where the delineation of teeth boundaries appears inaccurate, as evident in samples b, c, and d. These visual observations provide additional insights into the strengths and weaknesses of the competing methods.

%

%

\section{Conclusion}
The 3D Teeth Scan Segmentation and Labeling Challenge (3DTeethSeg'22) was conducted in conjunction with MICCAI 2022. The challenge provided a publicly available dataset consisting of 1800 intra-oral 3D scans obtained from 900 patients, which is currently the largest and most accurately annotated intra-oral 3D scans dataset. The challenge aimed to evaluate six algorithms for teeth detection, segmentation, and labeling tasks. This paper presents an overview of the challenge setup, summarizes the participating algorithms, and compares their performance. 
During the final testing phase of the 3D Teeth Scan Segmentation and Labeling Challenge (3DTeethSeg'22), the CGIP team's algorithm achieved the highest overall score of 0.9539, making it the top-performing solution. Additionally, the same algorithm obtained the highest performance in the segmentation task, with a score of 0.9859 for the TSA metric. On the other hand, the FiboSeg team's algorithms excelled in the teeth detection task, securing the top position with a score of 0.9924 for the Exp(-TLA) metric. For the labeling task, the IGIP team's algorithm achieved the highest score, attaining a value of 0.9289 for the TIR metric.

Future directions could include the incorporation of more variabilities in the dataset, such as more challenging cases with missing or damaged teeth and ambiguous labeling scenarios to provide a more comprehensive evaluation and enhance the algorithms' capability to handle real-world scenarios effectively. Additionally, it is important to evaluate the accuracy and smoothness of the predicted gum/teeth boundaries in the next iteration of the challenge. In future iterations, evaluating the run-time and computational complexity of the algorithms would be beneficial, as these factors are key considerations for integrating the developed solutions into Computer-Aided Design (CAD) software used in dentistry.

\bibliography{main}

\begin{thebibliography}{49}
\expandafter\ifx\csname natexlab\endcsname\relax\def\natexlab#1{#1}\fi
\providecommand{\url}[1]{\texttt{#1}}
\providecommand{\href}[2]{#2}
\providecommand{\path}[1]{#1}
\providecommand{\DOIprefix}{doi:}
\providecommand{\ArXivprefix}{arXiv:}
\providecommand{\URLprefix}{URL: }
\providecommand{\Pubmedprefix}{pmid:}
\providecommand{\doi}[1]{\href{http://dx.doi.org/#1}{\path{#1}}}
\providecommand{\Pubmed}[1]{\href{pmid:#1}{\path{#1}}}
\providecommand{\bibinfo}[2]{#2}
\ifx\xfnm\relax \def\xfnm[#1]{\unskip,\space#1}\fi
\bibitem[{Cui et~al.(2020)Cui, Li, Chen, Wei, Chen, Zhou \&
  Wang}]{cui2020tsegnet}
\bibinfo{author}{Cui, Z.}, \bibinfo{author}{Li, C.}, \bibinfo{author}{Chen,
  N.}, \bibinfo{author}{Wei, G.}, \bibinfo{author}{Chen, R.},
  \bibinfo{author}{Zhou, Y.}, \& \bibinfo{author}{Wang, W.}
  (\bibinfo{year}{2020}).
\newblock \bibinfo{title}{Tsegnet: an efficient and accurate tooth segmentation
  network on 3d dental model}.
\newblock {\it \bibinfo{journal}{Medical Image Analysis}\/},  {\it
  \bibinfo{volume}{69}\/}, \bibinfo{pages}{101949}.
\bibitem[{Cui et~al.(2019)Cui, Li \& Wang}]{cui2019toothnet}
\bibinfo{author}{Cui, Z.}, \bibinfo{author}{Li, C.}, \& \bibinfo{author}{Wang,
  W.} (\bibinfo{year}{2019}).
\newblock \bibinfo{title}{Toothnet: Automatic tooth instance segmentation and
  identification from cone beam ct images}.
\newblock In {\it \bibinfo{booktitle}{Proceedings of the IEEE/CVF Conference on
  Computer Vision and Pattern Recognition}\/} (pp.
  \bibinfo{pages}{6368--6377}).
\bibitem[{Eck et~al.(1995)Eck, DeRose, Duchamp, Hoppe, Lounsbery \&
  Stuetzle}]{eck1995multiresolution}
\bibinfo{author}{Eck, M.}, \bibinfo{author}{DeRose, T.},
  \bibinfo{author}{Duchamp, T.}, \bibinfo{author}{Hoppe, H.},
  \bibinfo{author}{Lounsbery, M.}, \& \bibinfo{author}{Stuetzle, W.}
  (\bibinfo{year}{1995}).
\newblock \bibinfo{title}{Multiresolution analysis of arbitrary meshes}.
\newblock In {\it \bibinfo{booktitle}{ACM Conference on Computer graphics and
  interactive techniques (SIGGRAPH '95)}\/} (pp. \bibinfo{pages}{173--182}).
\bibitem[{Ester et~al.(1996{\natexlab{a}})Ester, Kriegel, Sander \&
  Xu}]{Esteretal1996}
\bibinfo{author}{Ester, M.}, \bibinfo{author}{Kriegel, H.~P.},
  \bibinfo{author}{Sander, J.}, \& \bibinfo{author}{Xu, X.}
  (\bibinfo{year}{1996}{\natexlab{a}}).
\newblock \bibinfo{title}{A density-based algorithm for discovering clusters in
  large spatial databases with noise}.
\newblock In {\it \bibinfo{booktitle}{Proceedings of the Second International
  Conference on Knowledge Discovery and Data Mining}\/} (p.
  \bibinfo{pages}{226–231}).
\bibitem[{Ester et~al.(1996{\natexlab{b}})Ester, Kriegel, Sander \&
  Xu}]{dbscan}
\bibinfo{author}{Ester, M.}, \bibinfo{author}{Kriegel, H.-P.},
  \bibinfo{author}{Sander, J.}, \& \bibinfo{author}{Xu, X.}
  (\bibinfo{year}{1996}{\natexlab{b}}).
\newblock \bibinfo{title}{A density-based algorithm for discovering clusters in
  large spatial databases with noise}.
\newblock In {\it \bibinfo{booktitle}{Proceedings of the Second International
  Conference on Knowledge Discovery and Data Mining}\/} (pp.
  \bibinfo{pages}{226--–231}).
\bibitem[{Hatamizadeh et~al.(2022)Hatamizadeh, Tang, Nath, Yang, Myronenko,
  Landman, Roth \& Xu}]{hatamizadeh2022unetr}
\bibinfo{author}{Hatamizadeh, A.}, \bibinfo{author}{Tang, Y.},
  \bibinfo{author}{Nath, V.}, \bibinfo{author}{Yang, D.},
  \bibinfo{author}{Myronenko, A.}, \bibinfo{author}{Landman, B.},
  \bibinfo{author}{Roth, H.~R.}, \& \bibinfo{author}{Xu, D.}
  (\bibinfo{year}{2022}).
\newblock \bibinfo{title}{Unetr: Transformers for 3d medical image
  segmentation}.
\newblock In {\it \bibinfo{booktitle}{Proceedings of the IEEE/CVF winter
  conference on applications of computer vision}\/} (pp.
  \bibinfo{pages}{574--584}).
\bibitem[{{He} et~al.(2017){He}, {Gkioxari}, {Dollar} \& {Girshick}}]{8237584}
\bibinfo{author}{{He}, K.}, \bibinfo{author}{{Gkioxari}, G.},
  \bibinfo{author}{{Dollar}, P.}, \& \bibinfo{author}{{Girshick}, R.}
  (\bibinfo{year}{2017}).
\newblock \bibinfo{title}{Mask r-cnn}.
\newblock In {\it \bibinfo{booktitle}{2017 IEEE International Conference on
  Computer Vision (ICCV)}\/} (pp. \bibinfo{pages}{2980--2988}).
\bibitem[{Im et~al.(2022)Im, Kim, Yu, Lee, Choi, Kim, Ahn \&
  Cha}]{im2022accuracy}
\bibinfo{author}{Im, J.}, \bibinfo{author}{Kim, J.-Y.}, \bibinfo{author}{Yu,
  H.-S.}, \bibinfo{author}{Lee, K.-J.}, \bibinfo{author}{Choi, S.-H.},
  \bibinfo{author}{Kim, J.-H.}, \bibinfo{author}{Ahn, H.-K.}, \&
  \bibinfo{author}{Cha, J.-Y.} (\bibinfo{year}{2022}).
\newblock \bibinfo{title}{Accuracy and efficiency of automatic tooth
  segmentation in digital dental models using deep learning}.
\newblock {\it \bibinfo{journal}{Scientific Reports}\/},  {\it
  \bibinfo{volume}{12}\/}, \bibinfo{pages}{1--11}.
\bibitem[{Jiang et~al.(2020)Jiang, Zhao, Shi, Liu, Fu \& Jia}]{Jiangetal2020}
\bibinfo{author}{Jiang, L.}, \bibinfo{author}{Zhao, H.}, \bibinfo{author}{Shi,
  S.}, \bibinfo{author}{Liu, S.}, \bibinfo{author}{Fu, C.}, \&
  \bibinfo{author}{Jia, J.} (\bibinfo{year}{2020}).
\newblock \bibinfo{title}{Pointgroup: Dual-set point grouping for 3d instance
  segmentation}.
\newblock In {\it \bibinfo{booktitle}{Proceedings of the IEEE/CVF International
  Conference on Computer Vision}\/} (pp. \bibinfo{pages}{4866--4875}).
\bibitem[{Johnson et~al.(2019)Johnson, Douze \& J{\'e}gou}]{johnson2019billion}
\bibinfo{author}{Johnson, J.}, \bibinfo{author}{Douze, M.}, \&
  \bibinfo{author}{J{\'e}gou, H.} (\bibinfo{year}{2019}).
\newblock \bibinfo{title}{Billion-scale similarity search with gpus}.
\newblock {\it \bibinfo{journal}{IEEE Transactions on Big Data}\/},  {\it
  \bibinfo{volume}{7}\/}, \bibinfo{pages}{535--547}.
\bibitem[{Kronfeld et~al.(2010)Kronfeld, Brunner \&
  Brunnett}]{kronfeld2010snake}
\bibinfo{author}{Kronfeld, T.}, \bibinfo{author}{Brunner, D.}, \&
  \bibinfo{author}{Brunnett, G.} (\bibinfo{year}{2010}).
\newblock \bibinfo{title}{Snake-based segmentation of teeth from virtual dental
  casts}.
\newblock {\it \bibinfo{journal}{Computer-Aided Design and Applications}\/},
  {\it \bibinfo{volume}{7}\/}, \bibinfo{pages}{221--233}.
\bibitem[{Lai et~al.(2022)Lai, Liu, Jiang, Wang, Zhao, Liu, Qi \&
  Jia}]{stratified_transformer}
\bibinfo{author}{Lai, X.}, \bibinfo{author}{Liu, J.}, \bibinfo{author}{Jiang,
  L.}, \bibinfo{author}{Wang, L.}, \bibinfo{author}{Zhao, H.},
  \bibinfo{author}{Liu, S.}, \bibinfo{author}{Qi, X.}, \& \bibinfo{author}{Jia,
  J.} (\bibinfo{year}{2022}).
\newblock \bibinfo{title}{Stratified transformer for 3d point cloud
  segmentation}.
\bibitem[{Lian et~al.(2019)Lian, Wang, Wu, Liu, Dur{\'a}n, Ko \&
  Shen}]{lian2019meshsnet}
\bibinfo{author}{Lian, C.}, \bibinfo{author}{Wang, L.}, \bibinfo{author}{Wu,
  T.-H.}, \bibinfo{author}{Liu, M.}, \bibinfo{author}{Dur{\'a}n, F.},
  \bibinfo{author}{Ko, C.-C.}, \& \bibinfo{author}{Shen, D.}
  (\bibinfo{year}{2019}).
\newblock \bibinfo{title}{Meshsnet: Deep multi-scale mesh feature learning for
  end-to-end tooth labeling on 3d dental surfaces}.
\newblock In {\it \bibinfo{booktitle}{International Conference on Medical Image
  Computing and Computer-Assisted Intervention}\/} (pp.
  \bibinfo{pages}{837--845}).
\newblock \bibinfo{organization}{Springer}.
\bibitem[{Lian et~al.(2020)Lian, Wang, Wu, Wang, Yap, Ko \&
  Shen}]{lian2020deep}
\bibinfo{author}{Lian, C.}, \bibinfo{author}{Wang, L.}, \bibinfo{author}{Wu,
  T.-H.}, \bibinfo{author}{Wang, F.}, \bibinfo{author}{Yap, P.-T.},
  \bibinfo{author}{Ko, C.-C.}, \& \bibinfo{author}{Shen, D.}
  (\bibinfo{year}{2020}).
\newblock \bibinfo{title}{Deep multi-scale mesh feature learning for automated
  labeling of raw dental surfaces from 3d intraoral scanners}.
\newblock {\it \bibinfo{journal}{IEEE transactions on medical imaging}\/},
  {\it \bibinfo{volume}{39}\/}, \bibinfo{pages}{2440--2450}.
\bibitem[{Liao et~al.(2015)Liao, Liu, Zou, Ding, Liang \&
  Huang}]{liao2015automatic}
\bibinfo{author}{Liao, S.-h.}, \bibinfo{author}{Liu, S.-j.},
  \bibinfo{author}{Zou, B.-j.}, \bibinfo{author}{Ding, X.},
  \bibinfo{author}{Liang, Y.}, \& \bibinfo{author}{Huang, J.-h.}
  (\bibinfo{year}{2015}).
\newblock \bibinfo{title}{Automatic tooth segmentation of dental mesh based on
  harmonic fields}.
\newblock {\it \bibinfo{journal}{BioMed research international}\/},  {\it
  \bibinfo{volume}{2015}\/}.
\bibitem[{Liu et~al.(2021)Liu, Lin, Cao, Hu, Wei, Zhang, Lin \&
  Guo}]{swin_transformer}
\bibinfo{author}{Liu, Z.}, \bibinfo{author}{Lin, Y.}, \bibinfo{author}{Cao,
  Y.}, \bibinfo{author}{Hu, H.}, \bibinfo{author}{Wei, Y.},
  \bibinfo{author}{Zhang, Z.}, \bibinfo{author}{Lin, S.}, \&
  \bibinfo{author}{Guo, B.} (\bibinfo{year}{2021}).
\newblock \bibinfo{title}{Swin transformer: Hierarchical vision transformer
  using shifted windows}.
\bibitem[{Loshchilov \& Hutter(2017)}]{adamw}
\bibinfo{author}{Loshchilov, I.}, \& \bibinfo{author}{Hutter, F.}
  (\bibinfo{year}{2017}).
\newblock \bibinfo{title}{Decoupled weight decay regularization}.
\bibitem[{Ma et~al.(2020)Ma, Wei, Zhou, Pan, Xin \& Wang}]{ma2020srf}
\bibinfo{author}{Ma, Q.}, \bibinfo{author}{Wei, G.}, \bibinfo{author}{Zhou,
  Y.}, \bibinfo{author}{Pan, X.}, \bibinfo{author}{Xin, S.}, \&
  \bibinfo{author}{Wang, W.} (\bibinfo{year}{2020}).
\newblock \bibinfo{title}{{SRF}‐{Net}: {Spatial} {Relationship} {Feature}
  {Network} for {Tooth} {Point} {Cloud} {Classification}}.
\newblock {\it \bibinfo{journal}{Computer Graphics Forum}\/},  {\it
  \bibinfo{volume}{39}\/}, \bibinfo{pages}{267--277}.
\bibitem[{Miki et~al.(2017)Miki, Muramatsu, Hayashi, Zhou, Hara, Katsumata \&
  Fujita}]{miki2017classification}
\bibinfo{author}{Miki, Y.}, \bibinfo{author}{Muramatsu, C.},
  \bibinfo{author}{Hayashi, T.}, \bibinfo{author}{Zhou, X.},
  \bibinfo{author}{Hara, T.}, \bibinfo{author}{Katsumata, A.}, \&
  \bibinfo{author}{Fujita, H.} (\bibinfo{year}{2017}).
\newblock \bibinfo{title}{Classification of teeth in cone-beam ct using deep
  convolutional neural network}.
\newblock {\it \bibinfo{journal}{Computers in biology and medicine}\/},  {\it
  \bibinfo{volume}{80}\/}, \bibinfo{pages}{24--29}.
\bibitem[{Paszke et~al.(2019)Paszke, Gross, Massa, Lerer, Bradbury, Chanan,
  Killeen, Lin, Gimelshein, Antiga, Desmaison, Kopf, Yang, DeVito, Raison,
  Tejani, Chilamkurthy, Steiner, Fang, Bai \& Chintala}]{pytorch}
\bibinfo{author}{Paszke, A.}, \bibinfo{author}{Gross, S.},
  \bibinfo{author}{Massa, F.}, \bibinfo{author}{Lerer, A.},
  \bibinfo{author}{Bradbury, J.}, \bibinfo{author}{Chanan, G.},
  \bibinfo{author}{Killeen, T.}, \bibinfo{author}{Lin, Z.},
  \bibinfo{author}{Gimelshein, N.}, \bibinfo{author}{Antiga, L.},
  \bibinfo{author}{Desmaison, A.}, \bibinfo{author}{Kopf, A.},
  \bibinfo{author}{Yang, E.}, \bibinfo{author}{DeVito, Z.},
  \bibinfo{author}{Raison, M.}, \bibinfo{author}{Tejani, A.},
  \bibinfo{author}{Chilamkurthy, S.}, \bibinfo{author}{Steiner, B.},
  \bibinfo{author}{Fang, L.}, \bibinfo{author}{Bai, J.}, \&
  \bibinfo{author}{Chintala, S.} (\bibinfo{year}{2019}).
\newblock \bibinfo{title}{Pytorch: An imperative style, high-performance deep
  learning library}.
\newblock In {\it \bibinfo{booktitle}{Advances in Neural Information Processing
  Systems 32}\/} (pp. \bibinfo{pages}{8024--8035}).
\bibitem[{Qi et~al.(2017{\natexlab{a}})Qi, Su, Mo \& Guibas}]{PointNet2017}
\bibinfo{author}{Qi, C.~R.}, \bibinfo{author}{Su, H.}, \bibinfo{author}{Mo,
  K.}, \& \bibinfo{author}{Guibas, L.~J.} (\bibinfo{year}{2017}{\natexlab{a}}).
\newblock \bibinfo{title}{{PointNet}: {Deep} {Learning} on {Point} {Sets} for
  {3D} {Classification} and {Segmentation}}.
\newblock In {\it \bibinfo{booktitle}{Proceedings of the IEEE conference on
  computer vision and pattern recognition}\/} (pp. \bibinfo{pages}{652--660}).
\bibitem[{Qi et~al.(2017{\natexlab{b}})Qi, Yi, Su \& Guibas}]{qi2017pointnet++}
\bibinfo{author}{Qi, C.~R.}, \bibinfo{author}{Yi, L.}, \bibinfo{author}{Su,
  H.}, \& \bibinfo{author}{Guibas, L.~J.} (\bibinfo{year}{2017}{\natexlab{b}}).
\newblock \bibinfo{title}{Pointnet++: Deep hierarchical feature learning on
  point sets in a metric space}.
\newblock {\it \bibinfo{journal}{Advances in neural information processing
  systems}\/},  {\it \bibinfo{volume}{30}\/}.
\bibitem[{Qiu et~al.(2022)Qiu, Ye, Chen, Liu, Han \& Cui}]{qiu2022darch}
\bibinfo{author}{Qiu, L.}, \bibinfo{author}{Ye, C.}, \bibinfo{author}{Chen,
  P.}, \bibinfo{author}{Liu, Y.}, \bibinfo{author}{Han, X.}, \&
  \bibinfo{author}{Cui, S.} (\bibinfo{year}{2022}).
\newblock \bibinfo{title}{Darch: Dental arch prior-assisted 3d tooth instance
  segmentation with weak annotations}.
\newblock In {\it \bibinfo{booktitle}{Proceedings of the IEEE/CVF Conference on
  Computer Vision and Pattern Recognition}\/} (pp.
  \bibinfo{pages}{20752--20761}).
\bibitem[{Rao et~al.(2020)Rao, Wang, Meng, Pu, Sun \& Wang}]{rao2020symmetric}
\bibinfo{author}{Rao, Y.}, \bibinfo{author}{Wang, Y.}, \bibinfo{author}{Meng,
  F.}, \bibinfo{author}{Pu, J.}, \bibinfo{author}{Sun, J.}, \&
  \bibinfo{author}{Wang, Q.} (\bibinfo{year}{2020}).
\newblock \bibinfo{title}{A symmetric fully convolutional residual network with
  dcrf for accurate tooth segmentation}.
\newblock {\it \bibinfo{journal}{IEEE Access}\/},  {\it \bibinfo{volume}{8}\/},
  \bibinfo{pages}{92028--92038}.
\bibitem[{Ren et~al.(2015)Ren, He, Girshick \& Sun}]{ren2015faster}
\bibinfo{author}{Ren, S.}, \bibinfo{author}{He, K.}, \bibinfo{author}{Girshick,
  R.}, \& \bibinfo{author}{Sun, J.} (\bibinfo{year}{2015}).
\newblock \bibinfo{title}{Faster r-cnn: Towards real-time object detection with
  region proposal networks}.
\newblock {\it \bibinfo{journal}{arXiv preprint arXiv:1506.01497}\/}, .
\bibitem[{Rodriguez \& Laio(2014)}]{alex2014cfdp}
\bibinfo{author}{Rodriguez, A.}, \& \bibinfo{author}{Laio, A.}
  (\bibinfo{year}{2014}).
\newblock \bibinfo{title}{Clustering by fast search and find of density peaks}.
\newblock {\it \bibinfo{journal}{Science}\/},  {\it \bibinfo{volume}{344}\/},
  \bibinfo{pages}{1492--1496}.
\bibitem[{Schubert et~al.(2017)Schubert, Sander, Ester, Kriegel \&
  Xu}]{schubert2017dbscan}
\bibinfo{author}{Schubert, E.}, \bibinfo{author}{Sander, J.},
  \bibinfo{author}{Ester, M.}, \bibinfo{author}{Kriegel, H.~P.}, \&
  \bibinfo{author}{Xu, X.} (\bibinfo{year}{2017}).
\newblock \bibinfo{title}{Dbscan revisited, revisited: why and how you should
  (still) use dbscan}.
\newblock {\it \bibinfo{journal}{ACM Transactions on Database Systems
  (TODS)}\/},  {\it \bibinfo{volume}{42}\/}, \bibinfo{pages}{1--21}.
\bibitem[{Sinthanayothin \& Tharanont(2008)}]{sinth}
\bibinfo{author}{Sinthanayothin, C.}, \& \bibinfo{author}{Tharanont, W.}
  (\bibinfo{year}{2008}).
\newblock \bibinfo{title}{Orthodontics treatment simulation by teeth
  segmentation and setup}.
\newblock In {\it \bibinfo{booktitle}{International Conference on Electrical
  Engineering/Electronics, Computer, Telecommunications and Information
  Technology (ECTICON'08)}\/} (pp. \bibinfo{pages}{81--84}).
\newblock \bibinfo{organization}{IEEE} volume~\bibinfo{volume}{1}.
\bibitem[{Sun et~al.(2020)Sun, Pei, Li, Song, Guo, Zha \&
  Xu}]{sun2020automatic}
\bibinfo{author}{Sun, D.}, \bibinfo{author}{Pei, Y.}, \bibinfo{author}{Li, P.},
  \bibinfo{author}{Song, G.}, \bibinfo{author}{Guo, Y.}, \bibinfo{author}{Zha,
  H.}, \& \bibinfo{author}{Xu, T.} (\bibinfo{year}{2020}).
\newblock \bibinfo{title}{Automatic tooth segmentation and dense correspondence
  of 3d dental model}.
\newblock In {\it \bibinfo{booktitle}{International Conference on Medical Image
  Computing and Computer-Assisted Intervention}\/} (pp.
  \bibinfo{pages}{703--712}).
\newblock \bibinfo{organization}{Springer}.
\bibitem[{{Sun} et~al.(2020){Sun}, {Pei}, {Song}, {Guo}, {Ma}, {Xu} \&
  {Zha}}]{sun2020tooth}
\bibinfo{author}{{Sun}, D.}, \bibinfo{author}{{Pei}, Y.},
  \bibinfo{author}{{Song}, G.}, \bibinfo{author}{{Guo}, Y.},
  \bibinfo{author}{{Ma}, G.}, \bibinfo{author}{{Xu}, T.}, \&
  \bibinfo{author}{{Zha}, H.} (\bibinfo{year}{2020}).
\newblock \bibinfo{title}{Tooth segmentation and labeling from digital dental
  casts}.
\newblock In {\it \bibinfo{booktitle}{IEEE International Symposium on
  Biomedical Imaging (ISBI'20)}\/}.
\bibitem[{Szymański \& Kajdanowicz(2017)}]{scikit_multilearn}
\bibinfo{author}{Szymański, P.}, \& \bibinfo{author}{Kajdanowicz, T.}
  (\bibinfo{year}{2017}).
\newblock \bibinfo{title}{A scikit-based python environment for performing
  multi-label classification}.
\bibitem[{Tang et~al.(2022)Tang, Zhan, Chen, Yu \& Tao}]{Tangetal2022}
\bibinfo{author}{Tang, L.}, \bibinfo{author}{Zhan, Y.}, \bibinfo{author}{Chen,
  Z.}, \bibinfo{author}{Yu, B.}, \& \bibinfo{author}{Tao, D.}
  (\bibinfo{year}{2022}).
\newblock \bibinfo{title}{Contrastive boundary learning for point cloud
  segmentation}.
\newblock In {\it \bibinfo{booktitle}{Proceedings of the IEEE/CVF International
  Conference on Computer Vision}\/} (pp. \bibinfo{pages}{8489--8499}).
\bibitem[{Tian et~al.(2019)Tian, Dai, Zhang, Yuan, Yu \&
  Cheng}]{tian2019automatic}
\bibinfo{author}{Tian, S.}, \bibinfo{author}{Dai, N.}, \bibinfo{author}{Zhang,
  B.}, \bibinfo{author}{Yuan, F.}, \bibinfo{author}{Yu, Q.}, \&
  \bibinfo{author}{Cheng, X.} (\bibinfo{year}{2019}).
\newblock \bibinfo{title}{Automatic classification and segmentation of teeth on
  3d dental model using hierarchical deep learning networks}.
\newblock {\it \bibinfo{journal}{IEEE Access}\/},  {\it \bibinfo{volume}{7}\/},
  \bibinfo{pages}{84817--84828}.
\bibitem[{Verma et~al.(2018)Verma, Boyer \&
  Verbeek}]{vermaFeaStNetFeatureSteeredGraph2018}
\bibinfo{author}{Verma, N.}, \bibinfo{author}{Boyer, E.}, \&
  \bibinfo{author}{Verbeek, J.} (\bibinfo{year}{2018}).
\newblock \bibinfo{title}{{FeaStNet}: {Feature}-{Steered} {Graph}
  {Convolutions} for {3D} {Shape} {Analysis}}.
\newblock In {\it \bibinfo{booktitle}{IEEE Conference on {Computer} {Vision}
  and {Pattern} {Recognition} (CVPR)}\/} (pp. \bibinfo{pages}{2598--2606}).
\bibitem[{Wang et~al.(2020)Wang, Sun, Cheng, Jiang, Deng, Zhao, Liu, Mu, Tan,
  Wang et~al.}]{wang2020deep}
\bibinfo{author}{Wang, J.}, \bibinfo{author}{Sun, K.}, \bibinfo{author}{Cheng,
  T.}, \bibinfo{author}{Jiang, B.}, \bibinfo{author}{Deng, C.},
  \bibinfo{author}{Zhao, Y.}, \bibinfo{author}{Liu, D.}, \bibinfo{author}{Mu,
  Y.}, \bibinfo{author}{Tan, M.}, \bibinfo{author}{Wang, X.} et~al.
  (\bibinfo{year}{2020}).
\newblock \bibinfo{title}{Deep high-resolution representation learning for
  visual recognition}.
\newblock {\it \bibinfo{journal}{IEEE transactions on pattern analysis and
  machine intelligence}\/},  {\it \bibinfo{volume}{43}\/},
  \bibinfo{pages}{3349--3364}.
\bibitem[{Wang et~al.(2016)Wang, Yang, Mao, Huang, Huang \& Xu}]{wang2016cnn}
\bibinfo{author}{Wang, J.}, \bibinfo{author}{Yang, Y.}, \bibinfo{author}{Mao,
  J.}, \bibinfo{author}{Huang, Z.}, \bibinfo{author}{Huang, C.}, \&
  \bibinfo{author}{Xu, W.} (\bibinfo{year}{2016}).
\newblock \bibinfo{title}{Cnn-rnn: A unified framework for multi-label image
  classification}.
\newblock In {\it \bibinfo{booktitle}{IEEE conference on Computer Vision and
  Pattern Recognition}\/} (pp. \bibinfo{pages}{2285--2294}).
\bibitem[{Wu et~al.(2014)Wu, Chen, Li \& Zhou}]{wu2014tooth}
\bibinfo{author}{Wu, K.}, \bibinfo{author}{Chen, L.}, \bibinfo{author}{Li, J.},
  \& \bibinfo{author}{Zhou, Y.} (\bibinfo{year}{2014}).
\newblock \bibinfo{title}{Tooth segmentation on dental meshes using morphologic
  skeleton}.
\newblock {\it \bibinfo{journal}{Computers \& Graphics}\/},  {\it
  \bibinfo{volume}{38}\/}, \bibinfo{pages}{199--211}.
\bibitem[{Wu et~al.(2022)Wu, Lian, Lee, Pastewait, Piers, Liu, Wang, Wang,
  Chiu, Wang et~al.}]{wu2022two}
\bibinfo{author}{Wu, T.-H.}, \bibinfo{author}{Lian, C.}, \bibinfo{author}{Lee,
  S.}, \bibinfo{author}{Pastewait, M.}, \bibinfo{author}{Piers, C.},
  \bibinfo{author}{Liu, J.}, \bibinfo{author}{Wang, F.}, \bibinfo{author}{Wang,
  L.}, \bibinfo{author}{Chiu, C.-Y.}, \bibinfo{author}{Wang, W.} et~al.
  (\bibinfo{year}{2022}).
\newblock \bibinfo{title}{Two-stage mesh deep learning for automated tooth
  segmentation and landmark localization on 3d intraoral scans}.
\newblock {\it \bibinfo{journal}{IEEE Transactions on Medical Imaging}\/}, .
\bibitem[{Xu et~al.(2018)Xu, Liu \& Zheng}]{Xuetal2018}
\bibinfo{author}{Xu, X.}, \bibinfo{author}{Liu, C.}, \& \bibinfo{author}{Zheng,
  Y.} (\bibinfo{year}{2018}).
\newblock \bibinfo{title}{3d tooth segmentation and labeling using deep
  convolutional neural networks}.
\newblock {\it \bibinfo{journal}{IEEE transactions on visualization and
  computer graphics}\/},  {\it \bibinfo{volume}{25}\/},
  \bibinfo{pages}{2336--2348}.
\bibitem[{Yaqi \& Zhongke(2010)}]{yaqi2010computer}
\bibinfo{author}{Yaqi, M.}, \& \bibinfo{author}{Zhongke, L.}
  (\bibinfo{year}{2010}).
\newblock \bibinfo{title}{Computer aided orthodontics treatment by virtual
  segmentation and adjustment}.
\newblock In {\it \bibinfo{booktitle}{IEEE International Conference on Image
  Analysis and Signal Processing (ICISP)}\/} (pp. \bibinfo{pages}{336--339}).
\bibitem[{Yuan et~al.(2010)Yuan, Liao, Dai, Cheng \& Yu}]{old_ios_seg}
\bibinfo{author}{Yuan, T.}, \bibinfo{author}{Liao, W.}, \bibinfo{author}{Dai,
  N.}, \bibinfo{author}{Cheng, X.}, \& \bibinfo{author}{Yu, Q.}
  (\bibinfo{year}{2010}).
\newblock \bibinfo{title}{Single-tooth modeling for 3d dental model}.
\newblock {\it \bibinfo{journal}{International journal of biomedical
  imaging}\/},  {\it \bibinfo{volume}{2010}\/}.
\bibitem[{Zanjani et~al.(2019{\natexlab{a}})Zanjani, Moin, Claessen, Cherici,
  Parinussa, Pourtaherian \& Zinger}]{Zanjanietal2019}
\bibinfo{author}{Zanjani, F.~G.}, \bibinfo{author}{Moin, D.~A.},
  \bibinfo{author}{Claessen, F.}, \bibinfo{author}{Cherici, T.},
  \bibinfo{author}{Parinussa, S.}, \bibinfo{author}{Pourtaherian, A.}, \&
  \bibinfo{author}{Zinger, S.} (\bibinfo{year}{2019}{\natexlab{a}}).
\newblock \bibinfo{title}{Mask-mcnet: Instance segmentation in 3d point cloud
  of intra-oral scans}, .
\newblock (pp. \bibinfo{pages}{128--136}).
\bibitem[{Zanjani et~al.(2019{\natexlab{b}})Zanjani, Moin, Verheij, Claessen,
  Cherici, Tan et~al.}]{zanjani2019deep}
\bibinfo{author}{Zanjani, F.~G.}, \bibinfo{author}{Moin, D.~A.},
  \bibinfo{author}{Verheij, B.}, \bibinfo{author}{Claessen, F.},
  \bibinfo{author}{Cherici, T.}, \bibinfo{author}{Tan, T.} et~al.
  (\bibinfo{year}{2019}{\natexlab{b}}).
\newblock \bibinfo{title}{{Deep Learning Approach to Semantic Segmentation in
  3D Point Cloud Intra-oral Scans of Teeth}}.
\newblock In {\it \bibinfo{booktitle}{International Conference on Medical
  Imaging with Deep Learning (MIDL)}\/}.
\bibitem[{Zhang et~al.(2020)Zhang, Li, Song, Gao \& Lai}]{zhang2020automatic}
\bibinfo{author}{Zhang, J.}, \bibinfo{author}{Li, C.}, \bibinfo{author}{Song,
  Q.}, \bibinfo{author}{Gao, L.}, \& \bibinfo{author}{Lai, Y.-K.}
  (\bibinfo{year}{2020}).
\newblock \bibinfo{title}{{Automatic 3D Tooth Segmentation using Convolutional
  Neural Networks in Harmonic Parameter Space}}.
\newblock {\it \bibinfo{journal}{Elsevier Graphical Models}\/},  {\it
  \bibinfo{volume}{39}\/}, \bibinfo{pages}{101071}.
\bibitem[{Zhao et~al.(2021{\natexlab{a}})Zhao, Jiang, Jia, Torr \&
  Koltun}]{Zhaoetal2021}
\bibinfo{author}{Zhao, H.}, \bibinfo{author}{Jiang, L.}, \bibinfo{author}{Jia,
  J.}, \bibinfo{author}{Torr, P.~H.}, \& \bibinfo{author}{Koltun, V.}
  (\bibinfo{year}{2021}{\natexlab{a}}).
\newblock \bibinfo{title}{Point transformer}.
\newblock In {\it \bibinfo{booktitle}{Proceedings of the IEEE/CVF International
  Conference on Computer Vision}\/} (pp. \bibinfo{pages}{16259--16268}).
\bibitem[{Zhao et~al.(2006)Zhao, Ma, Tan \& Nie}]{zhao2006interactive}
\bibinfo{author}{Zhao, M.}, \bibinfo{author}{Ma, L.}, \bibinfo{author}{Tan,
  W.}, \& \bibinfo{author}{Nie, D.} (\bibinfo{year}{2006}).
\newblock \bibinfo{title}{Interactive tooth segmentation of dental models}.
\newblock In {\it \bibinfo{booktitle}{IEEE Engineering in Medicine and Biology
  Conference (IMBC'06)}\/} (pp. \bibinfo{pages}{654--657}).
\newblock \bibinfo{organization}{IEEE}.
\bibitem[{Zhao et~al.(2022)Zhao, Zhang, Liu, Meng, Cui, Gao, Gao, Lian \&
  Shen}]{zhao_two-stream_2022}
\bibinfo{author}{Zhao, Y.}, \bibinfo{author}{Zhang, L.}, \bibinfo{author}{Liu,
  Y.}, \bibinfo{author}{Meng, D.}, \bibinfo{author}{Cui, Z.},
  \bibinfo{author}{Gao, C.}, \bibinfo{author}{Gao, X.}, \bibinfo{author}{Lian,
  C.}, \& \bibinfo{author}{Shen, D.} (\bibinfo{year}{2022}).
\newblock \bibinfo{title}{Two-stream graph convolutional network for intra-oral
  scanner image segmentation}.
\newblock {\it \bibinfo{journal}{{IEEE} Transactions on Medical Imaging}\/},
  {\it \bibinfo{volume}{41}\/}.
\bibitem[{Zhao et~al.(2021{\natexlab{b}})Zhao, Zhang, Yang, Tan, Liu, Li, Huang
  \& Gao}]{zhao_3d_2021}
\bibinfo{author}{Zhao, Y.}, \bibinfo{author}{Zhang, L.}, \bibinfo{author}{Yang,
  C.}, \bibinfo{author}{Tan, Y.}, \bibinfo{author}{Liu, Y.},
  \bibinfo{author}{Li, P.}, \bibinfo{author}{Huang, T.}, \&
  \bibinfo{author}{Gao, C.} (\bibinfo{year}{2021}{\natexlab{b}}).
\newblock \bibinfo{title}{3d dental model segmentation with graph attentional
  convolution network}.
\newblock {\it \bibinfo{journal}{Pattern Recognition Letters}\/},  {\it
  \bibinfo{volume}{152}\/}, \bibinfo{pages}{79--85}.
\bibitem[{Zou et~al.(2015)Zou, Liu, Liao, Ding \& Liang}]{zou2015interactive}
\bibinfo{author}{Zou, B.-j.}, \bibinfo{author}{Liu, S.-j.},
  \bibinfo{author}{Liao, S.-h.}, \bibinfo{author}{Ding, X.}, \&
  \bibinfo{author}{Liang, Y.} (\bibinfo{year}{2015}).
\newblock \bibinfo{title}{Interactive tooth partition of dental mesh base on
  tooth-target harmonic field}.
\newblock {\it \bibinfo{journal}{Computers in biology and medicine}\/},  {\it
  \bibinfo{volume}{56}\/}, \bibinfo{pages}{132--144}.

\end{thebibliography}
\end{document}